\documentclass{article}
\usepackage{arxiv}
\usepackage[utf8]{inputenc} 
\usepackage[T1]{fontenc}    
\usepackage{natbib}         
\usepackage{hyperref}       
\usepackage{url}            
\usepackage{booktabs}       
\usepackage{amsfonts}       
\usepackage{nicefrac}       
\usepackage{microtype}      
\usepackage{xcolor}         
\usepackage{amsmath}
\usepackage{multirow}
\usepackage{makecell}
\usepackage{graphicx}
\usepackage{subcaption}
\usepackage{wrapfig}
\usepackage{rotating}

\title{Benchmarking Optimizers to Solve Inverse Problems \\ with Differentiable Physics Simulators}

\author{%
  Xiang Chen \qquad Huanhuan Xia \\[0.4em]
  Zhongguancun Academy, Beijing, China \\
  Zhongguancun Institute of Artificial Intelligence, Beijing, China \\[0.4em]
  \texttt{chenxiang@bza.edu.cn} \qquad
  \texttt{huanhuanxia@zgci.ac.cn}
}

\renewcommand{\shorttitle}{Benchmarking Optimizers for Differentiable Physics Inverse Problems}

\begin{document}

\maketitle

\begin{abstract}
Solving inverse problems with differentiable physics simulators holds the potential to revolutionize scientific discovery and engineering design, as it enjoys both the strict physical correctness from rigorous numerical physics simulators, and the high efficiency and effectiveness from automatic differentiation and gradient-based optimization. However, currently, this paradigm faces performance issues in optimization. In this work, we target benchmarking the performance of different optimizers to solve various inverse problems. We construct 12 differentiable physics simulators spanning physics domains including discrete mechanics, continuous mechanics, atomistic simulations, rendering, and semi-empirical physics models. Based on these simulators, we design corresponding inverse problems that can be categorized into parameter identification, inverse design, and optimal control. Finally, we conduct extensive experiments to compare the performance of different optimizers, including regular first-order methods, approximate second-order methods, as well as global optimizers, on these inverse problems, and analyze the results to provide insights on how to choose and design optimizers for differentiable programming. We hope such benchmarks can inspire the development of more effective optimizers, and further promote the applications of differentiable programming in various scientific and engineering domains.
\end{abstract}

\section{Introduction}
Deep learning has achieved remarkable success in the last decade, relying on the universal approximation capability of neural network models \citep{scarselli1998universal}, the high efficiency of gradient computation of automatic differentiation systems \citep{baydin2018automatic}, and the incredible effectiveness and generalizability from the optimization process \citep{chizat2020implicit}. However, purely data-driven models can never solve the issue of unguaranteed validity. One of the alternatives is differentiable physics models, which guarantee strict physical rigorousness by implementing the numerical physics process in automatic differentiation systems, while still enjoys the high efficiency and effectiveness provided by automatic differentiation and gradient-based optimizers.

Currently, differentiable physics models are facing performance issues. Although it achieves great success in certain scenarios, in some others the optimization process tends to fail. As a matter of fact, optimization in deep learning has never been an easy task. During the development of deep learning, many model architecture designs are specifically proposed so that training can be more effective and stable, such as ReLU \citep{nair2010rectified}, residual connections \citep{he2016deep}, normalizations \citep{huang2023normalization}, etc. However, such techniques can not be applied to differentiable physics models, because the model structure is predetermined by the underlying physical process. Hence this approach can only be improved by advancing optimizer techniques.

To the best of our knowledge, there are no comprehensive benchmarks available for evaluating the performance of optimizers on various differentiable physics models and inverse problems. Many previous works have implemented differentiable physics models for all kinds of physical domains and tasks in the last few years, but very few works have systematically studied this paradigm as a whole. It is difficult to unify these simulators and tasks into a framework, as each of them may use different coding structures, different automatic differentiation tools, and even different languages, which induces difficulties to discover better methods that can be widely applied in various scenarios. The closest previous works we can find are DiffTaichi \citep{Hu2020DiffTaichi} and PhiFlow \citep{holl2024phiflow}. However, their motivation and target are neither to benchmark optimizers nor to systematically study the performances on various physics models and inverse problems.

In this work, we target benchmarking the performance of different optimizers to solve inverse problems with differentiable physics simulators. We construct 12 differentiable physics simulators spanning areas including discrete mechanics, continuous mechanics, atomistic simulations, rendering, and semi-empirical physics models. Correspondingly, we design inverse problems for each of the physics simulators. The inverse problems are categorized into parameter identification, inverse design and optimal control problems. For optimizers, we utilize and implement first-order local, approximate second-order local, and global optimizers, each with 4 optimizers, all of them utilizing the gradient information to accelerate the optimization efficiency. Finally, we conduct extensive experiments to compare the performance of different optimizers on these problems, and analyze the results to provide insights on how to choose and design optimizers for differentiable programming. We hope such benchmarks can help make discoveries that are more fundamental and general about the optimization process of differentiable programming, inspire the development of more effective optimizers, and further promote the applications of differentiable programming in various scientific and engineering domains, as well as help general deep learning and large language model techniques to evolve to better model architectures and optimization methods.

\section{Related Work}

\paragraph{Differentiable Programming Techniques.}
Common deep learning layers such as fully-connected layers, convolution layers, and attention layers, are all explicit layers, which means the output of the layer $z$ can be written as an explicit function of the input $x$, parameterized by trainable parameter $\theta$, i.e., $z=f_{\theta}(x)$. Automatic differentiation, different from numerical differentiation or symbolic differentiation, can compute machine-precision gradients efficiently for models with high-dimensional input and low-dimensional output by backpropagation \citep{baydin2018automatic}.

For implicit layers, on the contrary, the input $x$ and output $z$ of the layer satisfy an equation $g_{\theta}(x,z)=0$ \citep{kolter2020implicit}. In such cases, the gradient can be computed by the implicit differentiation method. Previous works deriving and implementing implicit layers have been proposed in various forms, including ODE (Ordinary Differential Equations) \citep{chen2018neural}, PDE (Partial Differential Equations) \citep{mitusch2019dolfin}, self-consistent iterations \citep{bai2019deep}, optimization \citep{amos2017optnet,agrawal2019differentiable}, etc.

\paragraph{Differentiable Physics Models.} Over the past decade, differentiable physics models have been developed across a broad spectrum of physical domains. In classical mechanics, differentiable formulations have been extensively advanced for both discrete and continuous systems. For discrete mechanics, researchers have successfully differentiated rigid-body dynamics \citep{de2018end}, mass-spring systems \citep{Hu2020DiffTaichi}, and formalisms grounded in Lagrangian \citep{lutter2018deep} and Hamiltonian mechanics \citep{greydanus2019hamiltonian}. Parallel progress in continuous mechanics has enabled end-to-end differentiable simulations of fluid mechanics \citep{holl2024phiflow}, elastic mechanics \citep{hoyer2019neural}, as well as wave propagation in electromagnetics \citep{oskooi2010meep}, optics \citep{colburn2021inverse}, and acoustics \citep{jin2024diffsound}. Beyond classical regimes, differentiability has been systematically integrated into ab-initio simulations, including DFT (Density Functional Theory) \citep{li2021kohn,PhysRevLett.127.126403}, MD (Molecular Dynamics) \citep{doerr2021torchmd}, and high-accuracy quantum chemical methods such as CCSD/CI \citep{tamayo2018automatic}. In computer graphics and vision, these techniques have further catalyzed advances in inverse rendering \citep{kato2020differentiable}, such as NeRF (Neural Radiance Fields) \citep{mildenhall2021nerf} and 3D Gaussian Splatting \citep{kerbl20233d}. Underpinning this rapid interdisciplinary expansion are dedicated programming ecosystems and frameworks explicitly engineered for differentiable programming, most notably JAX \citep{schoenholz2020jax,freeman2021brax,xue2023jax,campagne2023jax,du2025jax}, Julia \citep{innes2019differentiable}, and Taichi \citep{Hu2020DiffTaichi}.

\paragraph{Optimizers for Deep Learning.} Standard deep learning optimizers are predominantly first-order local optimizers with mini-batch gradient estimation, such as SGD (Stochastic Gradient Descent) and its variants with momentum and adaptive learning rate \citep{sun2020optimization,schneider2019deepobs}. Recently, novel methods such as Muon \citep{jordan2024muon,liu2025muon} have achieved substantial improvement in training large language models, by explicitly leveraging the two-dimensional structure of neural network parameters. Concurrently, approximate second-order optimizers have been actively investigated for deep learning, including L-BFGS \citep{bollapragada2018progressive}, K-FAC \citep{martens2015optimizing}, Shampoo \citep{gupta2018shampoo}, and SOAP \citep{vyas2025soap}.

\section{Method}
In the following, we will introduce separately the differentiable physics models we implemented, the inverse problems we designed, and the optimizers we benchmarked in this work.

\subsection{Differentiable Physics Model Zoo}
\begin{table}[tb]
  \caption{List of the implemented differentiable physics models.}
  \label{table:models}
  \centering
  \begin{tabular}{llll}
    \toprule
    \textbf{Category} & \textbf{Model} & \textbf{Input} & \textbf{Output} \\
    \midrule
    \multirow{4}{*}{\makecell[l]{Discrete \\ Mechanics}} & Projectile Motion & \makecell[l]{Earth's gravity \\ initial state} & horizontal range \\
         \cmidrule(l){2-4}
         & Flight Dynamics        & \makecell[l]{vehicle params \\ Earth params \\ initial state \\ control policy} & flight trajectory \\
         \cmidrule(l){2-4}
         & Mass-Spring Systems    & \makecell[l]{system params \\ initial state \\ control policy} & movement trajectory \\
         \cmidrule(l){2-4}
         & Rigid-Body Dynamics    & \makecell[l]{system params \\ initial state \\ control policy} & movement trajectory \\
    \midrule
    \multirow{2}{*}{\makecell[l]{Continuous \\ Mechanics}} & Elastic Mechanics & \makecell[l]{material density \\ boundary conditions \\ physical params} & \makecell[l]{compliance \\ displacement} \\
         \cmidrule(l){2-4}
         & Fluid Mechanics & \makecell[l]{car shape \\ physical params} & drag \\
    \midrule
    \multirow{2}{*}{\makecell[l]{Atomistic \\ Simulations}} & Density Functional Theory & system params & \makecell[l]{total energy \\ electron density} \\
         \cmidrule(l){2-4}
         & Tight Binding & material params & \makecell[l]{eigenvalue spectrum \\ Bloch bands} \\
    \midrule
    \multirow{2}{*}{\makecell[l]{Rendering}} & Gaussian Splatting & \makecell[l]{scene params \\ render params} & scene image \\
         \cmidrule(l){2-4}
         & Atmospheric Scattering & \makecell[l]{Earth params \\ Sun params \\ atmosphere params \\ render params} & sky image \\
    \midrule
    \multirow{2}{*}{\makecell[l]{Semi-Empirical}} & Tire Magic Formula & tire params & tire forces \\
         \cmidrule(l){2-4}
         & Battery Thevenin Model & \makecell[l]{circuit params \\ OCV params \\ initial state \\ current profile} & \makecell[l]{voltage \\ SoC trajectory} \\
    \bottomrule
  \end{tabular}
\end{table}

\begin{figure}[htb]
  \centering
  \setkeys{Gin}{width=\textwidth,height=0.17\textheight,keepaspectratio}
  \begin{subfigure}[t]{0.32\textwidth}
    \centering
    \includegraphics{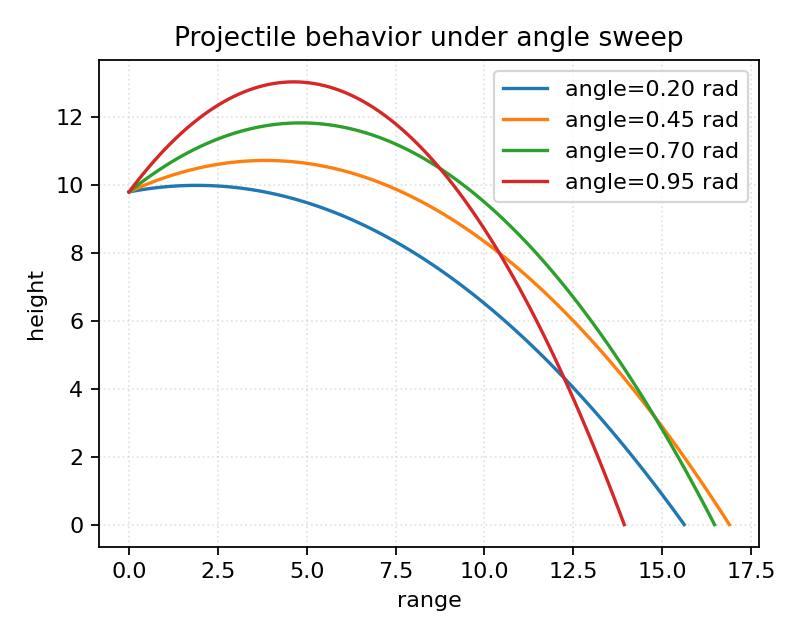}
    \caption{Projectile Motion}
  \end{subfigure}
  \hfill
  \begin{subfigure}[t]{0.32\textwidth}
    \centering
    \includegraphics{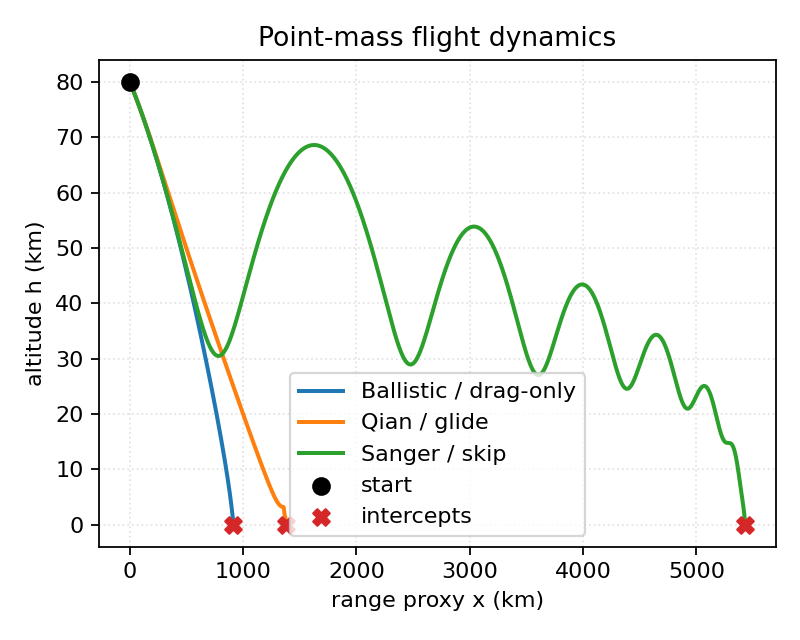}
    \caption{Flight Dynamics}
  \end{subfigure}
  \hfill
  \begin{subfigure}[t]{0.32\textwidth}
    \centering
    \includegraphics{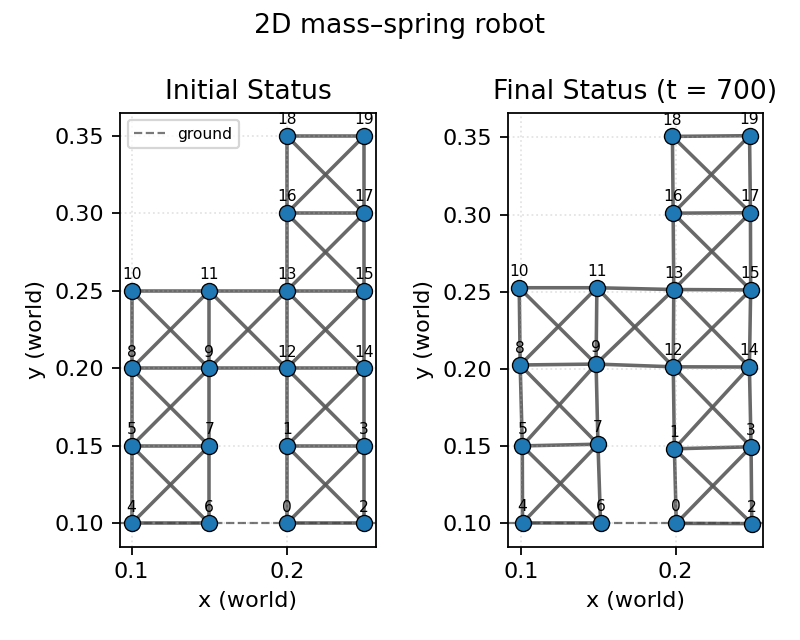}
    \caption{Mass-Spring Systems}
  \end{subfigure}
  \\
  \begin{subfigure}[t]{0.32\textwidth}
    \centering
    \includegraphics{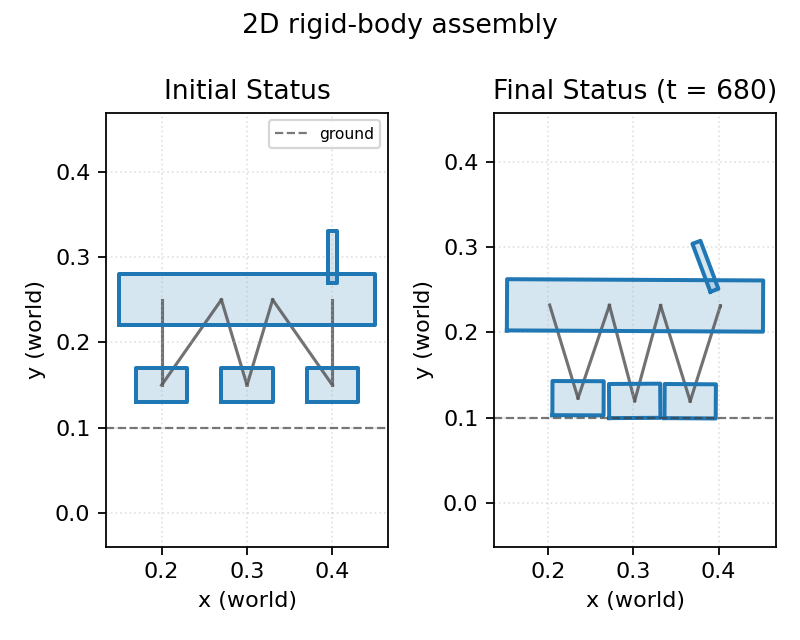}
    \caption{Rigid-Body Dynamics}
  \end{subfigure}
  \hfill
  \begin{subfigure}[t]{0.32\textwidth}
    \centering
    \includegraphics{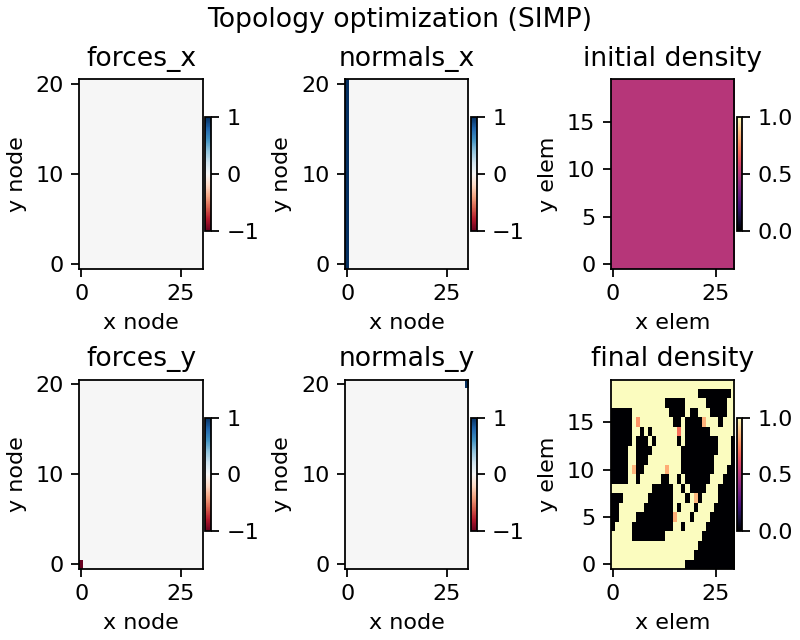}
    \caption{Elastic Mechanics}
  \end{subfigure}
  \hfill
  \begin{subfigure}[t]{0.32\textwidth}
    \centering
    \includegraphics{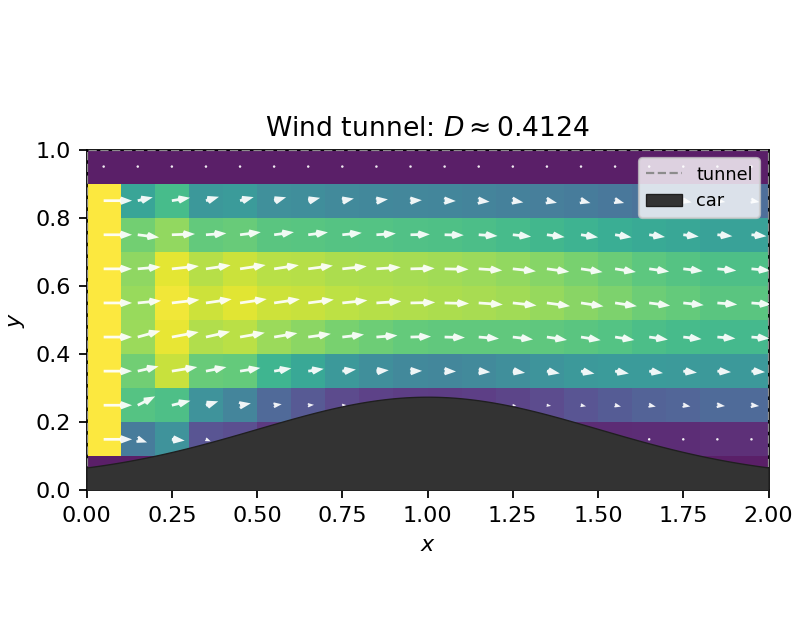}
    \caption{Fluid Mechanics}
  \end{subfigure}
  \\
  \begin{subfigure}[t]{0.32\textwidth}
    \centering
    \includegraphics{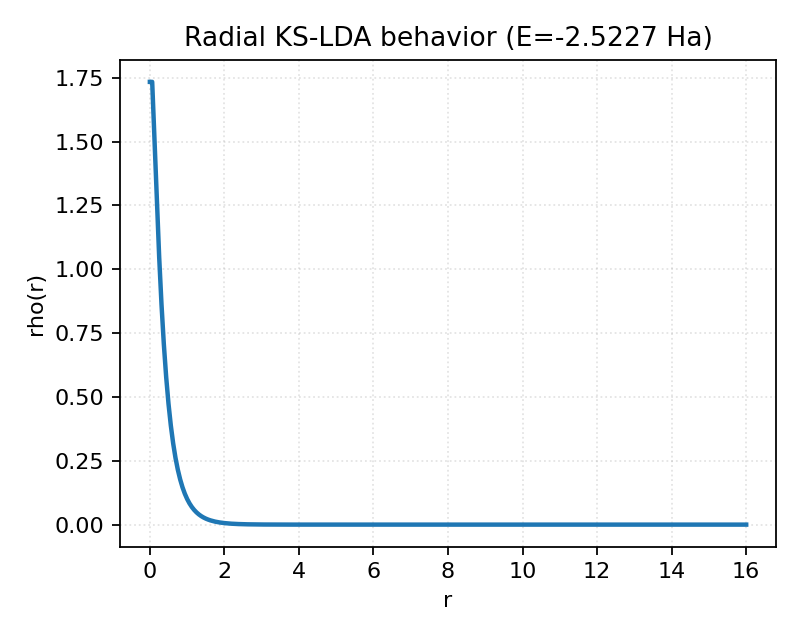}
    \caption{Density Functional Theory}
  \end{subfigure}
  \hfill
  \begin{subfigure}[t]{0.32\textwidth}
    \centering
    \includegraphics{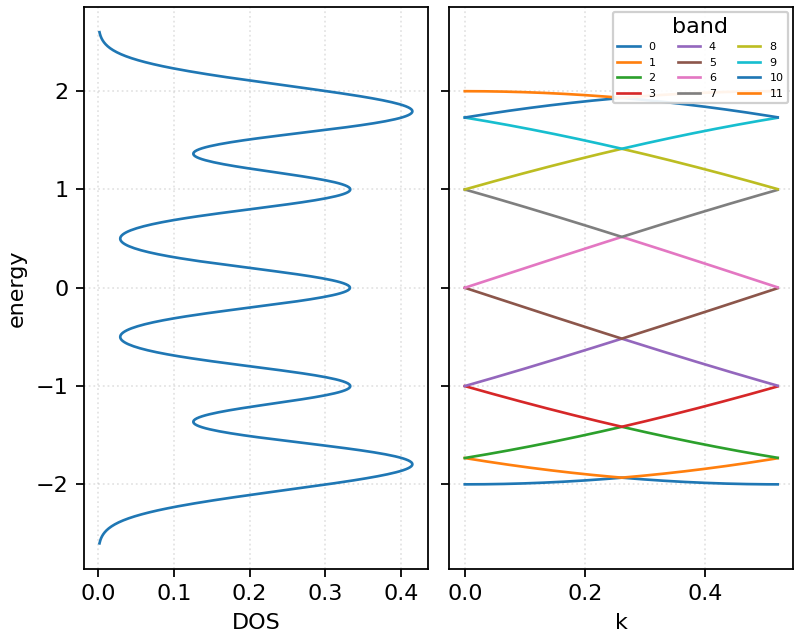}
    \caption{Tight Binding}
  \end{subfigure}
  \hfill
  \begin{subfigure}[t]{0.32\textwidth}
    \centering
    \includegraphics{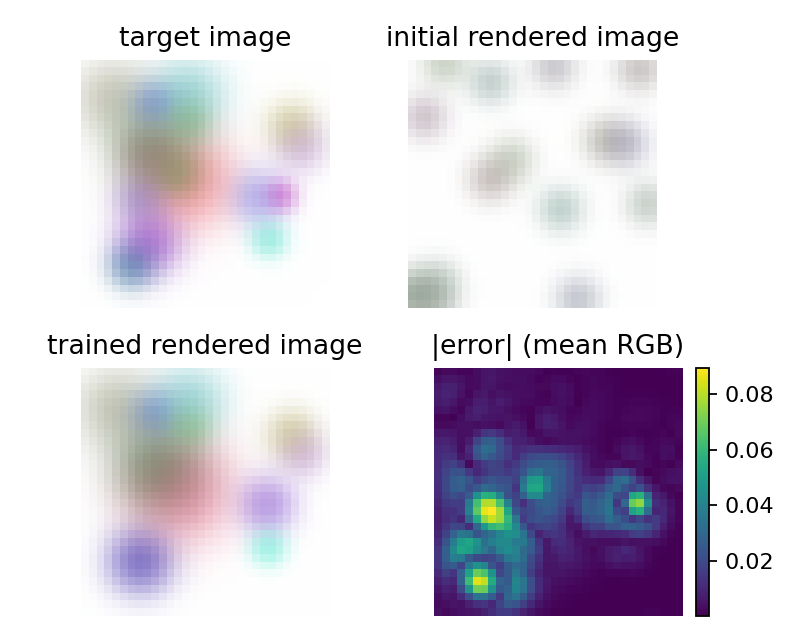}
    \caption{Gaussian Splatting}
  \end{subfigure}
  \\
  \begin{subfigure}[t]{0.32\textwidth}
    \centering
    \includegraphics{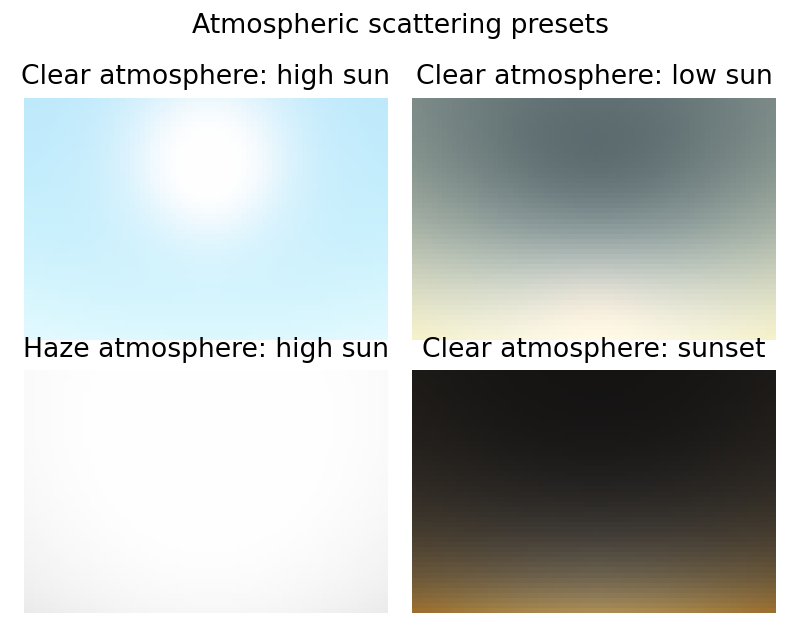}
    \caption{Atmospheric Scattering}
  \end{subfigure}
  \hfill
  \begin{subfigure}[t]{0.32\textwidth}
    \centering
    \includegraphics{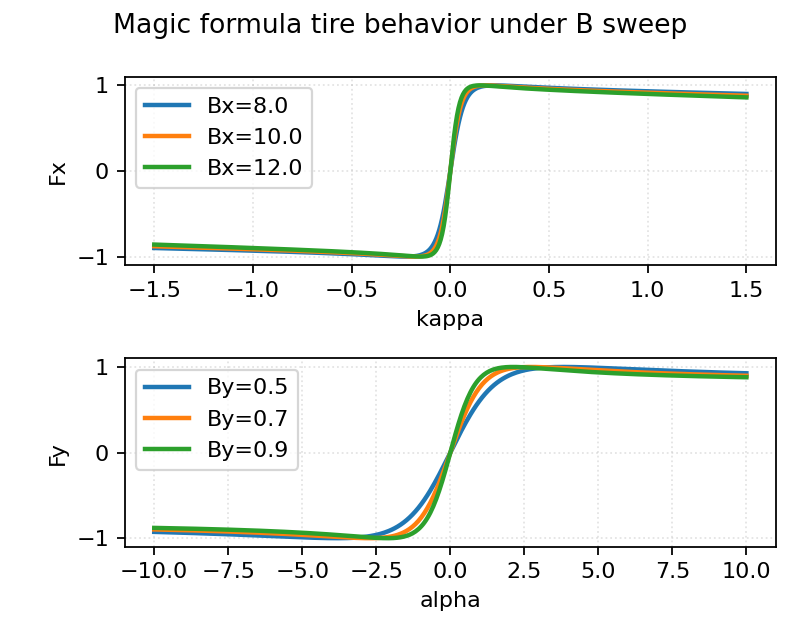}
    \caption{Tire Magic Formula}
  \end{subfigure}
  \hfill
  \begin{subfigure}[t]{0.32\textwidth}
    \centering
    \includegraphics{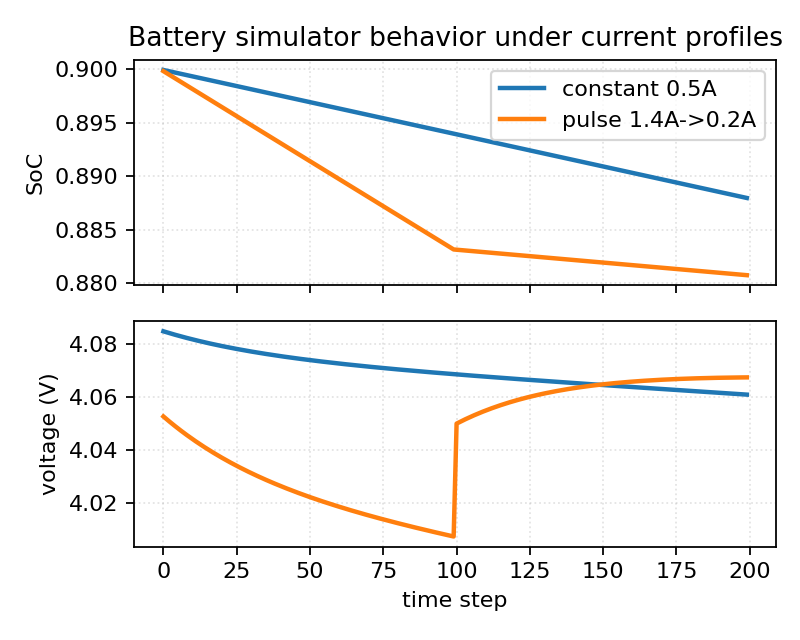}
    \caption{Battery Thevenin Model}
  \end{subfigure}
  \caption{Gallery of the implemented differentiable physics simulators.}
  \label{fig:simulator_gallery}
\end{figure}

The physics models we implemented in the model zoo are summarized in Table~\ref{table:models}. We consider 5 categories of physics simulations, i.e., discrete mechanics, continuous mechanics, atomistic simulations, rendering, and semi-empirical physics models, which can cover quite a portion of application scenarios such as academic research, industrial development, entertainment business, etc. All models are implemented in PyTorch, so that we enjoy a unified interface as well as the mature and active community support. All the physical and geometrical configurations, initial and boundary conditions, are provided in the \texttt{\_\_init\_\_} function. Only the control policies for controlled dynamic systems are fed from the \texttt{forward} function. The optimization variables are set as trainable, and the gradient of these variables with respect to the loss function is computed efficiently by the automatic differentiation system. We provide a gallery of the implemented differentiable physics simulators in Figure~\ref{fig:simulator_gallery} to demonstrate the sanity of our implementation. More implementation details, such as the underlying mathematical formulations and numerical algorithms of each simulator, are provided in Appendix~\ref{app:models}.

\paragraph{Discrete Mechanics}
The discrete mechanics simulators consider systems made up of discrete components. Within this category, we consider a simple projectile motion problem which, from specific initial height, speed, launch angle, and gravity, computes flight time and horizontal range; a flight dynamics model simulating the flight trajectory with atmosphere, gravity, and controllable lift/drag coefficients; a 2D mass-spring robot with springs, damping, gravity, and inelastic ground contact, driven by a control policy; and a controllable 2D rigid-body dynamics system with damped springs and joints.

\paragraph{Continuous Mechanics}
The continuous mechanics simulators target macroscopic continuum problems, which are typically governed by PDEs. In the current benchmark, we implemented an elastic mechanics simulator and a fluid simulator, both in 2 dimensions. The elastic mechanics simulator \citep{christensen2008introduction} computes the compliance and displacement of an elastic body under external forces and normals, given the material density and physical parameters. The fluid simulator computes the drag of a car silhouette in a wind tunnel \citep{angot1999penalization}.

\paragraph{Atomistic Simulations}
The atomistic simulations target microscopic problems at the atomic level. The DFT (Density Functional Theory) simulator \citep{bickelhaupt2000kohn} computes the total energy and electron density of a system of atoms, more specifically in our current implementation, a two-electron spherical closed shell. The TB (Tight Binding) model \citep{goringe1997tight} computes the supercell eigenvalue spectrum, Bloch bands, and (for visualization) a Gaussian-smoothed density of states, given onsite energies and hopping parameters.

\paragraph{Rendering}
The rendering simulations target image formation from environmental and rendering parameters. The Gaussian splatting simulator \citep{kerbl20233d} renders the image by splatting the gaussian blobs with properties including position, scale, and color. The atmospheric scattering simulator \citep{rayleigh1871scattering} generates the sky image by computing Rayleigh and Mie scattering, given the Earth, the Sun, and the atmosphere medium parameters.

\paragraph{Semi-Empirical Physics Models}
In many engineering scenarios, the underlying physical process is either not fully understood, or too complex to be efficiently computed. In such cases, semi-empirical physics models are used. We implement two popular semi-empirical models, namely the magic formula for tire dynamics \citep{pacejka2026magic}, and the Thevenin equivalent circuit model for battery SoC (State of Charge) estimation \citep{hannan2017review}.

\subsection{Inverse Problems}

We design inverse problems based on the constructed differentiable physics models, which can be categorized into 3 types, namely parameter identification, inverse design and optimal control. The dimensions of the optimization variables of different problems are designed to have different scales. We intentionally keep the problems small in scale, so that each optimizer can be benchmarked with all inverse problems in an efficient way. Also, we believe scaling law holds for the benchmarked problems, hence the performance on larger problems can be estimated from the results on small problems. The inverse problem tasks are summarized in Table~\ref{table:problems}.


\paragraph{Parameter Identification}
For parameter identification problems, the goal is to identify the unknown parameters in the physics model so that it best matches observed data.
For the XC Functional Fitting problem, we fit the LDA exchange-correlation functional to the reference total energy and electron density.
For the Tire Params Fitting problem, we fit the Pacejka BCDE parameters so longitudinal and lateral forces match synthetic curves from a hidden ground-truth.
For the Battery Params Fitting problem, we fit the first-order Thevenin circuit parameters together with a cubic open-circuit voltage map so that, under fixed random pulse-current profiles and initial state of charge, the predicted terminal voltage and state-of-charge trajectories match a frozen ground-truth battery, with voltage error as the primary term and a smaller weight on state-of-charge error.
For the Tight Binding Params Fitting problem, we fit onsite and hopping parameters so that the supercell eigenvalue spectrum and the Bloch band energies match a frozen reference chain, with a larger weight on the eigenvalues than on the bands; the density of states is computed for visualization but is not part of the training loss.
For the 2D Reconstruction problem, we fit the Gaussian blobs for the rendered image to match a target image.

\paragraph{Inverse Design}
For inverse design problems, the goal is to optimize certain parameters of the system so that a desired performance metric can achieve the best result.
The Range Matching problem of Projectile Motion fits the launch angle so that the predicted horizontal range matches a fixed target value, measured by squared error.
The Shape Optimization problem tunes the shape parameters of the car silhouette in order to minimize the drag estimate.
The Chromaticity Maximization problem is to discover the best sun direction and atmosphere medium parameters to maximize the RGB difference between the top and bottom of the rendered sky image.
The Topology Optimization problem optimizes element densities under the SIMP material law on a half MBB beam to minimize scaled compliance subject to a soft volume constraint that keeps the mean density near a prescribed volume fraction.

\paragraph{Optimal Control}
The optimal control problems are only for dynamic systems, where an optimal policy can be discovered for the system to achieve optimum of a desired performance.
The Range Maximization problem of Flight Dynamics learns an open-loop schedule of lift and drag coefficients over a fixed horizon. The simulator integrates altitude and ground range in meters, while the objective reports downrange in units of one thousand kilometers and adds small effort and temporal-smoothness penalties on the control logits.
The Robot Setpoint problem with Rigid-Body Dynamics learns open-loop spring actuations over a short horizon to drive a designated body toward a planar goal, with terminal velocity and control-regularization terms.
The Robot Reach problem with Mass-Spring Systems likewise learns open-loop spring actuations over a longer horizon to reach a planar goal, with analogous velocity and effort or smoothness penalties.

\begin{table}[tb]
  \caption{List of the benchmarked inverse problems.}
  \label{table:problems}
  \centering
  \begin{tabular}{llll}
    \toprule
    \textbf{Category} & \textbf{Problem} & \textbf{Base Simulator} & \textbf{\makecell[l]{Dimension of \\ Design Variables}} \\
    \midrule
    \multirow{5}{*}{\makecell[l]{Parameter \\ Identification}} & XC Functional Fitting & DFT & 2 \\
    & Tire Params Fitting & Tire Magic Formula & 8 \\
    & Battery Params Fitting & Battery Thevenin Model & 9 \\
    & Tight Binding Params Fitting & Tight Binding & 32 \\
    & 2D Reconstruction & Gaussian Splatting & 224 \\
    \midrule
    \multirow{4}{*}{\makecell[l]{Inverse \\ Design}}           & Range Matching & Projectile Motion & 1 \\
    & Shape Optimization & Fluid Mechanics & 4 \\
    & Chromaticity Maximization & Atmospheric Scattering & 15 \\
    & Topology Optimization & Elastic Mechanics & 600 \\
    \midrule
    \multirow{3}{*}{\makecell[l]{Optimal \\ Control}}          & Robot Setpoint & Rigid-Body Dynamics & 70 \\
    & Range Maximization & Flight Dynamics & 400 \\
    & Robot Reach & Mass-Spring Systems & 1100\\
    \bottomrule
  \end{tabular}
\end{table}

\subsection{Optimizers}
\begin{wraptable}{r}{0.5\textwidth}
  \caption{List of the benchmarked optimizers.}
  \label{table:optimizers}
  \centering
  \begin{tabular}{lll}
    \toprule
    \textbf{Category} & \textbf{Optimizer} \\
    \midrule
    \multirow{4}{*}{\makecell[l]{First-order \\ Local}} & SGD \\
         & RMSprop                \\
         & Adam                   \\
         & AdamW                  \\
    \midrule
    \multirow{4}{*}{\makecell[l]{Approximate \\ Second-order \\ Local}} & L-BFGS \\
         & K-FAC                  \\
         & Shampoo                \\
         & SOAP                   \\
    \midrule
    \multirow{4}{*}{\makecell[l]{Global}} & Random Search \\
         & Simulated Annealing    \\
         & Particle Swarm         \\
         & Differential Evolution \\
    \bottomrule
  \end{tabular}
\end{wraptable}

The optimizers we benchmark fall into 3 categories, namely first-order local optimizers, approximate second-order local optimizers, and global optimizers utilizing gradient information for local search. For each category, we implemented 4 optimizers, as listed in Table~\ref{table:optimizers}. For first-order local optimizers, they are the most commonly used optimizers in deep learning. In this work, we include PyTorch official implementations of SGD, RMSprop, Adam, and AdamW optimizers. For approximate second-order local optimizers, they have better convergence and accuracy properties than first-order optimizers, hence are suitable for scientific computing problems. Except for the classical L-BFGS method, the other 3 optimizers are modern variants that are specifically designed for deep learning, i.e., K-FAC \citep{martens2015optimizing}, Shampoo \citep{gupta2018shampoo} and SOAP \citep{vyas2025soap}. The above two categories of optimizers are all local optimizers, which means they are only capable of converging to local minima. However for problems with non-convex loss landscape and optimization variables with physical meanings, it is often desirable to find the global minimum. Hence we also include 4 well-recognized global optimizers, namely Random Search \citep{bergstra2012random}, Simulated Annealing \citep{kirkpatrick1983optimization}, Particle Swarm \citep{kennedy1995particle}, and Differential Evolution \citep{das2010differential}, which utilize gradient information for local search, but also have mechanisms to escape from local minima and explore the global landscape.

\section{Evaluation and Analysis}
In this section we benchmark the optimizers with the designed inverse problem tasks on the implemented differentiable physics models. We record the optimization process, time and memory consumption during experiments. Each experiment is repeated 3 times with random initialization to demonstrate statistical significance. All experiments are performed on a server with an NVIDIA A100 GPU (80\,GB). The major results are summarized in Table~\ref{table:results:params_ident}--\ref{table:results:optimal_control}, where each element represents the best objective values obtained during the optimization process, and the mean and std values are computed across the multiple runs of optimizations. For SIMP topology optimization, non-converged runs with negative best loss are excluded before hyper-parameter selection and aggregation (Appendix~\ref{app:experiment}). In the table, RS is short for Random Search, SA is short for Simulated Annealing, PS is short for Particle Swarm, and DE is short for Differential Evolution. More details regarding experimental setups and results are provided in Appendix~\ref{app:experiment}.

\begin{table}[t]
\caption{The performance of optimizers on the benchmarked parameter identification problems.}
\centering
\begin{tabular}{@{}lccccc@{}}
\toprule
Optimizer & DFT & Tire & Battery & Tight Binding & Gaussian Splatting \\
\midrule
SGD & $1.9\text{e}^{-14} \pm 2.7\text{e}^{-14}$ & $3.8\text{e}^{-4} \pm 1.9\text{e}^{-5}$ & $2.0\text{e}^{-5} \pm 1.4\text{e}^{-5}$ & $3.2\text{e}^{-5} \pm 2.1\text{e}^{-6}$ & $2.2\text{e}^{-2} \pm 4.7\text{e}^{-4}$ \\
RMSprop & $0.0 \pm 0.0$ & $2.6\text{e}^{-4} \pm 3.7\text{e}^{-5}$ & $2.5\text{e}^{-4} \pm 2.2\text{e}^{-4}$ & $5.6\text{e}^{-5} \pm 1.2\text{e}^{-5}$ & $1.6\text{e}^{-2} \pm 1.6\text{e}^{-4}$ \\
Adam & $1.9\text{e}^{-14} \pm 2.7\text{e}^{-14}$ & $2.6\text{e}^{-4} \pm 1.7\text{e}^{-5}$ & $5.1\text{e}^{-5} \pm 3.3\text{e}^{-5}$ & $7.2\text{e}^{-7} \pm 2.8\text{e}^{-7}$ & $1.4\text{e}^{-2} \pm 8.8\text{e}^{-5}$ \\
AdamW & $3.5\text{e}^{-12} \pm 3.0\text{e}^{-12}$ & $3.4\text{e}^{-4} \pm 1.5\text{e}^{-5}$ & $3.3\text{e}^{-4} \pm 3.6\text{e}^{-4}$ & $8.5\text{e}^{-7} \pm 4.9\text{e}^{-7}$ & $1.4\text{e}^{-2} \pm 1.6\text{e}^{-4}$ \\
LBFGS & $8.8\text{e}^{-11} \pm 6.2\text{e}^{-11}$ & $2.5\text{e}^{-7} \pm 4.0\text{e}^{-8}$ & $3.3\text{e}^{-6} \pm 5.2\text{e}^{-7}$ & $1.1\text{e}^{-9} \pm 5.2\text{e}^{-10}$ & $1.0\text{e}^{-2} \pm 1.9\text{e}^{-4}$ \\
KFAC & $5.1\text{e}^{-10} \pm 1.3\text{e}^{-10}$ & $2.4\text{e}^{-4} \pm 1.9\text{e}^{-5}$ & $4.9\text{e}^{-5} \pm 5.8\text{e}^{-5}$ & $3.5\text{e}^{-7} \pm 6.0\text{e}^{-8}$ & $2.2\text{e}^{-2} \pm 1.1\text{e}^{-3}$ \\
Shampoo & $0.0 \pm 0.0$ & $5.7\text{e}^{-4} \pm 1.0\text{e}^{-4}$ & $4.3\text{e}^{-4} \pm 5.4\text{e}^{-4}$ & $6.9\text{e}^{-4} \pm 2.2\text{e}^{-4}$ & $1.4\text{e}^{-2} \pm 1.0\text{e}^{-4}$ \\
SOAP & $3.2\text{e}^{-9} \pm 1.2\text{e}^{-9}$ & $2.5\text{e}^{-4} \pm 1.5\text{e}^{-5}$ & $5.6\text{e}^{-5} \pm 2.4\text{e}^{-5}$ & $4.1\text{e}^{-7} \pm 8.9\text{e}^{-8}$ & $1.5\text{e}^{-2} \pm 5.2\text{e}^{-4}$ \\
RS & $1.9\text{e}^{-14} \pm 2.7\text{e}^{-14}$ & $4.2\text{e}^{-4} \pm 3.2\text{e}^{-5}$ & $3.8\text{e}^{-5} \pm 1.1\text{e}^{-6}$ & $1.3\text{e}^{-4} \pm 1.6\text{e}^{-5}$ & $1.9\text{e}^{-2} \pm 3.4\text{e}^{-4}$ \\
SA & $9.3\text{e}^{-11} \pm 2.2\text{e}^{-11}$ & $1.1\text{e}^{-3} \pm 3.7\text{e}^{-4}$ & $4.5\text{e}^{-5} \pm 2.5\text{e}^{-6}$ & $5.2\text{e}^{-3} \pm 4.9\text{e}^{-4}$ & $2.2\text{e}^{-2} \pm 5.0\text{e}^{-4}$ \\
PS & $0.0 \pm 0.0$ & $8.1\text{e}^{-5} \pm 2.0\text{e}^{-5}$ & $8.5\text{e}^{-6} \pm 1.5\text{e}^{-6}$ & $2.0\text{e}^{-5} \pm 7.1\text{e}^{-6}$ & $1.9\text{e}^{-2} \pm 4.4\text{e}^{-4}$ \\
DE & $0.0 \pm 0.0$ & $1.3\text{e}^{-4} \pm 4.2\text{e}^{-5}$ & $1.7\text{e}^{-5} \pm 5.9\text{e}^{-6}$ & $4.4\text{e}^{-3} \pm 7.7\text{e}^{-4}$ & $2.6\text{e}^{-2} \pm 1.4\text{e}^{-3}$ \\
\bottomrule
\end{tabular}
\label{table:results:params_ident}
\end{table}

\begin{table}[t]
\caption{The performance of optimizers on the benchmarked inverse design problems.}
\centering
\begin{tabular}{@{}lcccc@{}}
\toprule
Optimizer & Projectile & Fluid & Atmospheric & Elastic \\
\midrule
SGD & $1.0\text{e}^{-9} \pm 1.3\text{e}^{-9}$ & $6.2\text{e}^{-2} \pm 4.6\text{e}^{-7}$ & $-7.5\text{e}^{-1} \pm 2.0\text{e}^{-3}$ & $1.5\text{e}^{0} \pm 4.8\text{e}^{-3}$ \\
RMSprop & $1.5\text{e}^{-9} \pm 1.9\text{e}^{-9}$ & $4.2\text{e}^{-2} \pm 1.7\text{e}^{-7}$ & $-7.5\text{e}^{-1} \pm 9.6\text{e}^{-5}$ & $1.6\text{e}^{0} \pm 5.5\text{e}^{-3}$ \\
Adam & $2.0\text{e}^{-10} \pm 1.5\text{e}^{-10}$ & $4.0\text{e}^{-2} \pm 1.3\text{e}^{-7}$ & $-7.5\text{e}^{-1} \pm 6.2\text{e}^{-4}$ & $1.5\text{e}^{0} \pm 5.2\text{e}^{-3}$ \\
AdamW & $2.9\text{e}^{-11} \pm 2.1\text{e}^{-11}$ & $4.2\text{e}^{-2} \pm 2.3\text{e}^{-7}$ & $-7.4\text{e}^{-1} \pm 1.9\text{e}^{-3}$ & $1.5\text{e}^{0} \pm 6.0\text{e}^{-3}$ \\
LBFGS & $2.4\text{e}^{-10} \pm 1.7\text{e}^{-10}$ & $4.0\text{e}^{-2} \pm 2.1\text{e}^{-8}$ & $-7.5\text{e}^{-1} \pm 4.9\text{e}^{-8}$ & $1.5\text{e}^{0} \pm 4.5\text{e}^{-3}$ \\
KFAC & $5.7\text{e}^{-10} \pm 1.9\text{e}^{-10}$ & $4.0\text{e}^{-2} \pm 9.8\text{e}^{-9}$ & $-7.5\text{e}^{-1} \pm 2.1\text{e}^{-5}$ & $1.6\text{e}^{0} \pm 1.2\text{e}^{-2}$ \\
Shampoo & $5.8\text{e}^{-7} \pm 5.1\text{e}^{-8}$ & $4.0\text{e}^{-2} \pm 0.0$ & $-7.5\text{e}^{-1} \pm 6.7\text{e}^{-4}$ & $1.6\text{e}^{0} \pm 8.7\text{e}^{-3}$ \\
SOAP & $6.2\text{e}^{-11} \pm 4.1\text{e}^{-11}$ & $4.0\text{e}^{-2} \pm 7.3\text{e}^{-6}$ & $-7.5\text{e}^{-1} \pm 4.8\text{e}^{-5}$ & $1.9\text{e}^{0} \pm 3.5\text{e}^{-2}$ \\
RS & $0.0 \pm 0.0$ & $4.0\text{e}^{-2} \pm 0.0$ & $-7.2\text{e}^{-1} \pm 4.0\text{e}^{-5}$ & $1.5\text{e}^{0} \pm 1.6\text{e}^{-2}$ \\
SA & $2.4\text{e}^{-12} \pm 1.7\text{e}^{-12}$ & $4.0\text{e}^{-2} \pm 0.0$ & $-7.2\text{e}^{-1} \pm 2.0\text{e}^{-2}$ & $2.6\text{e}^{0} \pm 5.7\text{e}^{-2}$ \\
PS & $0.0 \pm 0.0$ & $4.0\text{e}^{-2} \pm 0.0$ & $-7.5\text{e}^{-1} \pm 5.6\text{e}^{-8}$ & $1.5\text{e}^{0} \pm 6.6\text{e}^{-3}$ \\
DE & $0.0 \pm 0.0$ & $4.0\text{e}^{-2} \pm 1.8\text{e}^{-9}$ & $-7.5\text{e}^{-1} \pm 1.3\text{e}^{-4}$ & $2.5\text{e}^{0} \pm 2.0\text{e}^{-2}$ \\
\bottomrule
\end{tabular}
\label{table:results:inverse_design}
\end{table}

\begin{table}[t]
\caption{The performance of optimizers on the benchmarked optimal control problems.}
\centering
\begin{tabular}{@{}lccc@{}}
\toprule
Optimizer & Flight Dynamics & Rigid Body & Mass Spring \\
\midrule
SGD & $-1.3\text{e}^{0} \pm 1.2\text{e}^{-6}$ & $1.9\text{e}^{-1} \pm 1.7\text{e}^{-5}$ & $1.9\text{e}^{-1} \pm 1.1\text{e}^{-2}$ \\
RMSprop & $-1.3\text{e}^{0} \pm 6.7\text{e}^{-7}$ & $1.9\text{e}^{-1} \pm 1.3\text{e}^{-5}$ & $5.9\text{e}^{-2} \pm 9.6\text{e}^{-3}$ \\
Adam & $-1.3\text{e}^{0} \pm 9.7\text{e}^{-8}$ & $1.9\text{e}^{-1} \pm 2.7\text{e}^{-5}$ & $1.3\text{e}^{-1} \pm 1.3\text{e}^{-2}$ \\
AdamW & $-1.3\text{e}^{0} \pm 3.5\text{e}^{-7}$ & $1.9\text{e}^{-1} \pm 1.5\text{e}^{-5}$ & $1.4\text{e}^{-1} \pm 1.9\text{e}^{-2}$ \\
LBFGS & $-1.3\text{e}^{0} \pm 5.0\text{e}^{-6}$ & $1.9\text{e}^{-1} \pm 1.6\text{e}^{-3}$ & $2.0\text{e}^{-1} \pm 3.0\text{e}^{-2}$ \\
KFAC & $-1.3\text{e}^{0} \pm 5.6\text{e}^{-8}$ & $1.9\text{e}^{-1} \pm 7.8\text{e}^{-6}$ & $1.9\text{e}^{-1} \pm 1.0\text{e}^{-2}$ \\
Shampoo & $-1.3\text{e}^{0} \pm 0.0$ & $1.9\text{e}^{-1} \pm 2.5\text{e}^{-6}$ & $5.8\text{e}^{-2} \pm 1.0\text{e}^{-2}$ \\
SOAP & $-1.3\text{e}^{0} \pm 2.6\text{e}^{-3}$ & $1.9\text{e}^{-1} \pm 1.1\text{e}^{-5}$ & $1.1\text{e}^{-1} \pm 8.4\text{e}^{-3}$ \\
RS & $-1.3\text{e}^{0} \pm 2.2\text{e}^{-4}$ & $1.9\text{e}^{-1} \pm 4.1\text{e}^{-6}$ & $9.5\text{e}^{-2} \pm 1.6\text{e}^{-2}$ \\
SA & $-1.3\text{e}^{0} \pm 1.8\text{e}^{-3}$ & $1.9\text{e}^{-1} \pm 2.0\text{e}^{-5}$ & $1.1\text{e}^{-1} \pm 1.9\text{e}^{-2}$ \\
PS & $-1.3\text{e}^{0} \pm 1.1\text{e}^{-3}$ & $1.9\text{e}^{-1} \pm 3.6\text{e}^{-5}$ & $1.0\text{e}^{-1} \pm 3.6\text{e}^{-2}$ \\
DE & $-1.3\text{e}^{0} \pm 9.2\text{e}^{-4}$ & $1.9\text{e}^{-1} \pm 1.4\text{e}^{-4}$ & $1.1\text{e}^{-1} \pm 5.7\text{e}^{-3}$ \\
\bottomrule
\end{tabular}
\label{table:results:optimal_control}
\end{table}




\paragraph{When curvature helps, and when it hurts.}
Across parameter identification (Table~\ref{table:results:params_ident}), approximate second-order methods are not uniformly better, but they decisively shine on structured spectral matching: L-BFGS reaches a mean best loss on the order of $10^{-9}$ on tight-binding fitting, with K-FAC and SOAP close behind at $\sim\!10^{-7}$, whereas untuned first-order baselines such as SGD plateau several orders of magnitude higher. This matches the intuition that least-squares-type objectives against smooth matrix spectra are locally analytic and benefit from exploiting gradient geometry. Conversely, on the radial LDA exchange--correlation fit, nearly all optimizer families reach near-zero best losses (many within numerical tolerance of $0$) once hyperparameters are selected per method, indicating a well-conditioned two-parameter fit when the forward SCF remains stable. Tight-binding and DFT therefore bracket a key lesson for differentiable physics benchmarks: optimizer rankings are most informative on tasks where the forward model yields stable objectives but the landscape still separates update rules.

\paragraph{Global search, initialization sensitivity, and dimensionality.}
The expectation that global optimizers can be more robust under random initialization is supported most cleanly on low-dimensional inverse design. On projectile range matching (Table~\ref{table:results:inverse_design}), random search, particle swarm, and differential evolution reach essentially optimal best objectives (exact zeros within numerical tolerance), on par with L-BFGS/SOAP. Simulated annealing likewise attains a mean best loss of $\sim\!10^{-12}$, indicating repeated landing on the global structure of this one-dimensional landscape. At the same time, approximate second-order methods remain competitive here (L-BFGS mean $\sim\!10^{-10}$), illustrating the classical trade-off: when the basin is generous, curvature-aware updates converge explosively; when it is not, population-based methods pay extra forward passes but buy insurance against bad seeds.

The topology-optimization column is the counterpoint: with hundreds of design variables, most first-order methods and several global methods attain stable mean best losses around $1.5$, while simulated annealing and differential evolution plateau higher near $2.5$--$2.6$. Without a non-negativity filter, a minority of K-FAC and differential-evolution runs collapse to large negative losses (about $3\%$ of all SIMP runs, concentrated in these two optimizers); na\"ive minimization of mean best loss would then prefer those failed trajectories. After discarding non-converged runs ($\texttt{best\_loss}<0$) and re-selecting hyperparameters on converged repeats only, K-FAC lands with the stable cluster ($\sim\!1.6$), whereas differential evolution remains among the weaker but finite performers. Each compliance evaluation requires solving a large sparse equilibrium system inside the autodiff graph, so high-variance proposals can still probe physically meaningless density fields---but such failures should be filtered rather than rewarded. This supports the following conclusion that, global optimizers improve robustness primarily when either (i) the search space is low-dimensional, or (ii) stochastic gradient noise dominates and local curvature is uninformative---not automatically when the forward model is ``physics-based.''

\paragraph{Rendering and semi-empirical fits.}
Gaussian splatting reconstruction (224 parameters) shows a clear split: L-BFGS achieves the strongest mean best loss around $10^{-2}$, adaptive first-order methods land nearby, and global methods remain slightly worse around $2\times10^{-2}$ with small run-to-run variance. Semi-empirical tire and battery fits are comparatively forgiving: most methods land in a band of $10^{-6}$--$10^{-3}$ mean best losses, with L-BFGS particularly strong on both tasks and particle swarm competitive on the tire task. These tasks behave closer to ``standard'' differentiable programming in machine learning, where tuned first-order and second-order methods are strong baselines and globals serve as sanity checks rather than sole workhorses.

\paragraph{Optimal control: long unrolls and noisy credit assignment.}
The flight-dynamics and mass--spring columns (Table~\ref{table:results:optimal_control}) exemplify the point about backpropagating through long rollouts. Flight dynamics now yields tightly clustered mean best losses around $-1.3$ with very small standard deviations across all twelve optimizers, indicating a comparatively well-conditioned open-loop objective at the reported scale, with only modest separation among update rules. Mass--spring control remains more discriminative (mean best losses $\sim\!0.06$--$0.20$), with Shampoo and RMSprop among the strongest configurations, suggesting that even stable contact-rich rollouts retain multi-modal structure in the policy space. Rigid-body setpoint control in Table~\ref{table:results:optimal_control} attains finite mean best losses for all twelve optimizers, clustered around $\sim\!0.19$ with very small spread; this is qualitatively similar to the tight flight cluster and tighter than mass--spring, indicating a relatively stable setpoint objective here despite contacts and joints, while still leaving limited room for optimizer choice.

\paragraph{Compute--objective trade-offs.}
Appendix~\ref{app:compute_resources} makes the runtime story quantitative. On DFT XC fitting, a single global-optimizer step costs roughly $5$--$1.9\times10^{1}$\,s of wall time versus $\sim\!1$\,s for first- and second-order local methods---an order-of-magnitude gap before counting how many steps each family needs. The ratio is milder on inexpensive forward models (tire, Gaussian splatting) but widens again on elastic topology optimization and flight dynamics, where one global step can exceed ten forward--backward passes of a local optimizer. We therefore come to the following conclusion that, for offline calibration, design, or scientific inversion, paying an order of magnitude in wall time can be rational if it reduces residual error or avoids manual initialization engineering; for real-time control or any setting where the optimizer sits inside an inner loop at kHz rates, first-order methods with small per-step cost remain the pragmatic default, and the benchmark encourages reporting both objective and step-time tables whenever a new simulator is added.

\paragraph{Takeaway for future optimizer research.}
Differentiable physics does not inherit the ``default optimizers'' culture of deep networks: the same Adam hyperparameters that suffice for splatting can be less decisive for a Kohn--Sham outer loop, while L-BFGS may dominate on a 32-parameter tight-binding ring yet fail to offset the cost of global search on a 600-variable topology field. We hope the zoo encourages methods that adapt to simulator structure---mixed-precision stable eigen-solvers for quantum tasks, trust-region filters for contact rollouts, and hybrid global-local schedules whose search radius shrinks as physics residuals fall---rather than treating optimizer choice as an afterthought bolted onto an otherwise frozen forward model.

\section{Conclusion}
In this work, we constructed to the best of our knowledge the first benchmark for optimizers to solve inverse problems with differentiable physics simulators. More specifically, we implemented 12 differentiable physics simulators across various physics domains, and designed corresponding inverse problems spanning parameter identification, inverse design, and optimal control. For optimizers, we considered first-order, approximate second-order, and global optimizers, all utilizing the gradient information provided by the differentiable physics models and automatic differentiation tools. We benchmarked the performance of optimizers on these inverse problems, and analyzed the results to provide insights on how to choose and design optimizers for differentiable programming.

\paragraph{Limitations and Broader impacts.} For inverse problem task types, besides the 3 types we benchmarked in this work, there are also other types of tasks that can be solved by differentiable physics models, such as data assimilation and state estimation. Also, constrained optimization has not been considered in the current work, which is also ubiquitous in real-life applications. We will develop more physics simulators, task types and optimizers continuously. Broader impacts of this work include the potential to further promote the applications of differentiable programming in various scientific and engineering domains.

\subsubsection*{Acknowledgments}
This work was supported by Zhongguancun Academy under Grant No. XTS0030.

\bibliographystyle{unsrtnat}
\bibliography{references}


\appendix

\section{Details of Differentiable Physics Model Zoo}
\label{app:models}

\subsection{Discrete Mechanics}

\paragraph{Projectile Motion.}
We model planar motion of a point mass under constant gravitational acceleration $g>0$, with initial height $h_0>0$, launch speed $v_0>0$, and launch angle $\theta$. The initial velocity components are $v_{x,0}=v_0\cos\theta$ and $v_{y,0}=v_0\sin\theta$. The trajectory is the  parabola:
\begin{equation}
x(t)=v_{x,0}t,\qquad
y(t)=h_0+v_{y,0}t-\tfrac{1}{2}g t^2.
\end{equation}
The simulator is implemented in closed form: the time of flight to ground impact is obtained by solving $y(t)=0$ for the positive root,
\begin{equation}
t_\mathrm{imp}=\frac{v_{y,0}+\sqrt{v_{y,0}^2+2g h_0}}{g},
\end{equation}
and the horizontal range used as the scalar output is $s_x=v_{x,0}t_\mathrm{imp}$.

\paragraph{Flight Dynamics.}
The vehicle is modeled as a point mass $m$ with reference area $S$.
The system state is set as $\mathbf{s}=(h,x,v,\gamma)^\top$: altitude $h$ (m), a ground-range coordinate $x$ (m), speed $v$ (m/s), and flight-path angle $\gamma$ (rad, from horizontal). At each step, a control policy returns aerodynamic coefficients $(C_L,C_D)$ as functions of $\mathbf{s}$. Let $R_e$ denote the Earth radius and $\mu$ denote the gravitational parameter. The radial distance $r=R_e+h$ gives $g(r)=\mu/r^2$. The air density $\rho(h)=\rho_0\exp(-h/H)$ for $h\le 120\,\mathrm{km}$ and $\rho=0$ above. The dynamic pressure is $q=\tfrac{1}{2}\rho(h)v^2$, with lift $L=q S C_L$ and drag $D=q S C_D$. The system state is updated by:
\begin{align}
\dot h &= v\sin\gamma,\\
\dot x &= v\cos\gamma\,\frac{R_e}{r},\\
\dot v &= -\frac{D}{m}-g(r)\sin\gamma,\\
\dot\gamma &= \frac{L}{m v}-\left(\frac{g(r)}{v}-\frac{v}{r}\right)\cos\gamma.
\end{align}
One explicit Euler update with step $\Delta t$ advances $\mathbf{s}\leftarrow \mathbf{s}+\Delta t\,(\dot h,\dot x,\dot v,\dot\gamma)^\top$.
Rollouts start from a stored initial state and end until $h\le 0$.
In the benchmarked open-loop task, $(C_L,C_D)$ are produced from per-step logits as above; positions remain in SI meters inside the integrator, while the loss and trajectory plots divide $h$ and $x$ by $U=10^{6}\,\mathrm{m}$ so that the range term $-\,x_T/U$ is expressed in units of $1000\,\mathrm{km}$, together with the effort/smoothness regularizers scaled by $1/U$.

\paragraph{Mass-Spring Systems.}
The mathematical formulation and experimental setups are inspired by \citet[Sec. E.5]{Hu2020DiffTaichi}. We consider a two-dimensional particle system with $N$ point masses.
Let $\mathbf{x}_i(t),\,\mathbf{v}_i(t)\in\mathbb{R}^2$ denote the position and velocity of mass $i\in\{1,\ldots,N\}$.
Springs are undirected edges $e=(a,b)$ between anchors $a,b\in\{1,\ldots,N\}$.
Each spring has a nominal rest length $L_e>0$, a stiffness $k_e\ge 0$, and an actuation gain $\alpha_e\ge 0$.
A scalar control $u_e^n\in[-1,1]$ is applied at discrete time index $n$; passive springs correspond to $\alpha_e=0$, for which $u_e^n$ has no effect.
The commanded natural length at step $n$ is $L^{\mathrm{tar}}_e(u_e^n) \;=\; L_e\,\bigl(1 + \alpha_e\, u_e^n\bigr)$.
Let $\mathbf{d}^n_{e} \;=\; \mathbf{x}^n_a - \mathbf{x}^n_b$ and $\ell^n_e = \|\mathbf{d}^n_e\|$. For numerical stability we use a regularized length $\tilde{\ell}^n_e = \max(\ell^n_e,\varepsilon)$ with fixed small $\varepsilon>0$.
The spring exerts a velocity impulse aligned with $\mathbf{d}^n_e$:
\begin{equation}
  \boldsymbol{\Delta}\mathbf{v}^n_e
  \;=\;
  \Delta t\, k_e\,
  \frac{\tilde{\ell}^n_e - L^{\mathrm{tar}}_e(u_e^n)}{\tilde{\ell}^n_e}\,
  \mathbf{d}^n_e,
\end{equation}
which is accumulated on the endpoints as
$\mathbf{v}_a \leftarrow \mathbf{v}_a - \boldsymbol{\Delta}\mathbf{v}^n_e$ and
$\mathbf{v}_b \leftarrow \mathbf{v}_b + \boldsymbol{\Delta}\mathbf{v}^n_e$.

Let $\zeta>0$ be a damping coefficient and $g<0$ a constant gravitational acceleration along the vertical axis (so the gravitational increment to velocity is $(0,\,g\,\Delta t)$).
After summing spring impulses, velocities are updated by exponential damping and gravity:
\begin{equation}
  \widetilde{\mathbf{v}}^{\,n+1}_i
  \;=\;
  e^{-\zeta\,\Delta t}\,\mathbf{v}^n_i
  \;+\;
  \bigl(0,\,g\,\Delta t\bigr)
  \;+\;
  \sum_{e\ni i} \delta\mathbf{v}^n_{e,i},
\end{equation}
where $\delta\mathbf{v}^n_{e,i}$ is the contribution of spring $e$ to mass $i$ from the impulse rule above.

Let $y$ denote the vertical component and let the ground be the horizontal line $y=h$.
If the pre-update configuration is below the ground, $y^n_i<h$, and the post-impulse vertical velocity is still directed into the ground, $\widetilde{v}^{\,n+1}_{i,y}<0$, we apply an inelastic stop projection:
\begin{equation}
  \mathbf{v}^{n+1}_i \;=\;
  \begin{cases}
    \mathbf{0}, & y^n_i < h \;\;\text{and}\;\; \widetilde{v}^{\,n+1}_{i,y}<0,\\[0.25em]
    \widetilde{\mathbf{v}}^{\,n+1}_i, & \text{otherwise.}
  \end{cases}
\end{equation}
Positions are then advanced in semi-implicit Euler: $\mathbf{x}^{n+1}_i \;=\; \mathbf{x}^n_i + \Delta t\,\mathbf{v}^{n+1}_i$.

\paragraph{Rigid-Body Dynamics.}
The mathematical formulation and experimental setups are inspired by \citet[Sec. E.7]{Hu2020DiffTaichi}. The simulator tracks $N$ planar rigid bodies. For each body $i\in\{1,\ldots,N\}$ at discrete time $n$, let $\mathbf{x}_i^n\in\mathbb{R}^2$ be the center of mass, $\mathbf{v}_i^n\in\mathbb{R}^2$ the translational velocity, $\theta_i^n$ the orientation, and $\omega_i^n$ the scalar angular velocity (about the axis normal to the plane). Each body is a centered axis-aligned rectangle in its own frame with half-extents $(h_{x,i},h_{y,i})$. With uniform areal density $\rho=1$, the reference implementation uses
\begin{equation}
  m_i = 4\,h_{x,i}h_{y,i},
  \qquad
  I_i = \tfrac{4}{3}\,h_{x,i}h_{y,i}\,\bigl(h_{x,i}^2+h_{y,i}^2\bigr),
\end{equation}
with inverse parameters $m_i^{-1}$ and $I_i^{-1}$ used in impulse updates. Let $R(\theta)$ denote the plane rotation by $\theta$. A body-fixed attachment offset $\mathbf{r}_i^{\mathrm{b}}$ maps to world position $\mathbf{p}_i^n=\mathbf{x}_i^n+R(\theta_i^n)\mathbf{r}_i^{\mathrm{b}}$ and inherits linear velocity
\begin{equation}
  \mathbf{v}_{\mathrm{att},i}^n
  = \mathbf{v}_i^n + \omega_i^n\,\mathbf{J}\,R(\theta_i^n)\mathbf{r}_i^{\mathrm{b}},
  \qquad
  \mathbf{J}=\begin{bmatrix}0&-1\\[2pt]1&0\end{bmatrix}.
\end{equation}

Between bodies $a$ and $b$, two body-fixed offsets define world anchors $\mathbf{p}_a^n$ and $\mathbf{p}_b^n$ as above. Define $\mathbf{d}^n=\mathbf{p}_a^n-\mathbf{p}_b^n$ and the regularized length $\tilde{\ell}^n=\max\bigl(\|\mathbf{d}^n\|,\varepsilon\bigr)$ with fixed $\varepsilon=10^{-4}$. For a scalar control $u^n\in[-1,1]$ (clamped), a passive actuation gain $\alpha=0$ enforces an inactive control channel. For a spring with rest length $L>0$ and stiffness $k$, the commanded length is
\begin{equation}
  L_{\mathrm{tar}}^n(u^n)=L\,\bigl(1+\alpha\,u^n\bigr).
\end{equation}
For a joint, one sets $L_{\mathrm{tar}}^n=0$ so the anchors coincide in the unconstrained limit. The primary Hooke-style velocity impulse aligned with $\mathbf{d}^n$ is
\begin{equation}
  \mathbf{J}_{\mathrm{spring}}^n
  = \Delta t\,k\,
  \frac{\tilde{\ell}^n-L_{\mathrm{tar}}^n}{\tilde{\ell}^n}\,\mathbf{d}^n.
\end{equation}
For joints, an additional projection impulse damps the relative anchor velocity $\mathbf{v}_{\mathrm{rel}}^n=\mathbf{v}_{\mathrm{att},a}^n-\mathbf{v}_{\mathrm{att},b}^n$: with $\widehat{\mathbf{n}}_{\mathrm{rel}}^n=\mathbf{v}_{\mathrm{rel}}^n/\max\bigl(\|\mathbf{v}_{\mathrm{rel}}^n\|,\eta\bigr)$ and a small $\eta>0$, and with the usual effective inertia scalar $M_{\mathrm{eff}}^n$ assembled from $m_a^{-1}$, $m_b^{-1}$, $I_a^{-1}$, $I_b^{-1}$ and squared projected lever arms $\bigl(\widehat{\mathbf{n}}_{\mathrm{rel}}^n\times\mathbf{r}_a^n\bigr)^2$, $\bigl(\widehat{\mathbf{n}}_{\mathrm{rel}}^n\times\mathbf{r}_b^n\bigr)^2$ (two-dimensional scalar cross products), the joint contribution is
\begin{equation}
  \mathbf{J}_{\mathrm{joint}}^n
  = \frac{\|\mathbf{v}_{\mathrm{rel}}^n\|}{M_{\mathrm{eff}}^n}\,\widehat{\mathbf{n}}_{\mathrm{rel}}^n,
  \qquad
  \mathbf{J}^n = \mathbf{J}_{\mathrm{spring}}^n+\mathbf{J}_{\mathrm{joint}}^n.
\end{equation}
The link impulse is applied as equal-and-opposite linear impulses at the anchors: body $a$ receives $-\mathbf{J}^n$ and body $b$ receives $+\mathbf{J}^n$, each decomposed into center-of-mass increments
\begin{equation}
  \Delta \mathbf{v}_a^n = -\,\frac{\mathbf{J}^n}{m_a},
  \qquad
  \Delta \omega_a^n = \frac{(\mathbf{p}_a^n-\mathbf{x}_a^n)\times \mathbf{J}^n}{I_a},
\end{equation}
and the opposite signs for body $b$, with $\times$ denoting the scalar $z$-component of the planar cross product.

For each body, four corners are generated by independent sign patterns on $(\pm h_{x,i},\pm h_{y,i})$ in body frame. For each corner, let $\boldsymbol{\rho}_i^n$ be the world vector from the center of mass to the corner (after rotation), with corner position $\mathbf{c}_i^n=\mathbf{x}_i^n+\boldsymbol{\rho}_i^n$ and corner velocity $\mathbf{w}_i^n=\mathbf{v}_i^n+\omega_i^n\,\mathbf{J}\boldsymbol{\rho}_i^n$. For the ground test, the velocity is augmented by gravity over one step, $\mathbf{w}_i^{n,\ast}=\mathbf{w}_i^n+(0,g\Delta t)^\top$, and the predictor position is $\widehat{\mathbf{c}}_i^{\,n}=\mathbf{c}_i^n+\Delta t\,\mathbf{w}_i^{n,\ast}$. With horizontal ground $y=h$ and outward normal $\mathbf{n}=(0,1)^\top$, contact activation uses $\mathbf{n}^\top \mathbf{w}_i^{n,\ast}<0$ and $\widehat{c}_{i,y}^n<h$. The normal impulse magnitude $J_{n,i}^n\ge 0$ removes incoming normal motion in the inelastic limit with restitution coefficient $e$ (typically $e=0$),
\begin{equation}
  J_{n,i}^n
  = \max\!\left(0,\;
  -\frac{(1+e)\,\mathbf{n}^\top \mathbf{w}_i^{n,\ast}}
       {m_i^{-1} + (\mathbf{n}\times\boldsymbol{\rho}_i^n)^2 I_i^{-1}}\right),
\end{equation}
evaluated only when the contact predicate holds (otherwise $J_{n,i}^n=0$). With tangent $\mathbf{t}=(1,0)^\top$, a tangential impulse $J_{t,i}^n$ is first computed as if frictionless along $\mathbf{t}$, then Coulomb-capped,
\begin{equation}
  J_{t,i}^{n,\mathrm{raw}}
  = -\frac{\mathbf{t}^\top \mathbf{w}_i^{n,\ast}}
         {m_i^{-1} + (\mathbf{t}\times\boldsymbol{\rho}_i^n)^2 I_i^{-1}},
  \qquad
  J_{t,i}^n = \mathrm{clip}\!\bigl(J_{t,i}^{n,\mathrm{raw}},\,-\mu J_{n,i}^n,\,\mu J_{n,i}^n\bigr),
\end{equation}
with friction coefficient $\mu$. The world impulse is $\mathbf{I}_i^n=J_{n,i}^n\mathbf{n}+J_{t,i}^n\mathbf{t}$ and is mapped to rigid-body increments using the lever arm $\widehat{\mathbf{c}}_i^{\,n}-\mathbf{x}_i^n$ in the same cross-product formula as for links.

Let $\zeta$ be the exponential damping rate and $g$ the gravitational acceleration in the vertical direction. After summing all link and ground impulse contributions into $(\Delta\mathbf{v}_i^n,\Delta\omega_i^n)$, the velocity update is
\begin{equation}
  \widetilde{\mathbf{v}}_i^{\,n+1}
  = e^{-\zeta\Delta t}\,\mathbf{v}_i^n + (0,g\Delta t)^\top + \Delta\mathbf{v}_i^n,
  \qquad
  \widetilde{\omega}_i^{\,n+1}
  = e^{-\zeta\Delta t}\,\omega_i^n + \Delta\omega_i^n,
\end{equation}
followed by the semi-implicit position update
\begin{equation}
  \mathbf{x}_i^{n+1}=\mathbf{x}_i^n+\Delta t\,\widetilde{\mathbf{v}}_i^{\,n+1},
  \qquad
  \theta_i^{n+1}=\theta_i^n+\Delta t\,\widetilde{\omega}_i^{\,n+1}.
\end{equation}

\subsection{Continuous Mechanics}

\paragraph{Elastic Mechanics.} The mathematical formulation and experimental setups are inspired by \citet{hoyer2019neural}. We model small-deformation, plane-stress linear elasticity on a rectangular domain
$\Omega = (0,L_x)\times(0,L_y)$ discretized by a structured mesh of four-node quadrilateral (Q4)
elements with bilinear isoparametric shape functions $N_a(\xi,\eta)$, $a=1,\ldots,4$, on the
reference cell $(\xi,\eta)\in[-1,1]^2$.
Let $\mathcal{N}$ and $\mathcal{E}$ denote the sets of nodes and elements, and let
$\mathbf{u}\in\mathbb{R}^{2|\mathcal{N}|}$ collect nodal displacements $(u_x,u_y)$ in a fixed
global ordering.
Plane stress relates in-plane strains
$\boldsymbol{\varepsilon}=[\varepsilon_{xx},\varepsilon_{yy},\gamma_{xy}]^\top$
to stresses $\boldsymbol{\sigma}=[\sigma_{xx},\sigma_{yy},\tau_{xy}]^\top$ via
$\boldsymbol{\sigma}=\mathbf{D}(E,\nu)\,\boldsymbol{\varepsilon}$, where for isotropic modulus
$E$ and Poisson ratio $\nu$,
\begin{equation}
  \mathbf{D}(E,\nu)
  = \frac{E}{1-\nu^2}
  \begin{bmatrix}
    1 & \nu & 0 \\
    \nu & 1 & 0 \\
    0 & 0 & (1-\nu)/2
  \end{bmatrix}.
\end{equation}
For each element $e$, the strain–displacement matrix $\mathbf{B}_e(\mathbf{x})$ links element
nodal displacements $\mathbf{u}_e\in\mathbb{R}^8$ to strains
$\boldsymbol{\varepsilon}(\mathbf{x})=\mathbf{B}_e(\mathbf{x})\mathbf{u}_e$.
The element stiffness with uniform Young's modulus $E$ is
\begin{equation}
  \mathbf{K}_e(E)
  = \int_{\Omega_e} \mathbf{B}_e^\top \,\mathbf{D}(E,\nu)\, \mathbf{B}_e \,\mathrm{d}A,
\end{equation}
evaluated numerically with a $2\times2$ Gauss quadrature rule on $\Omega_e$.
The global stiffness assembles contributions from all elements,
$\mathbf{K}(\{E_e\})=\sum_e \mathbf{L}_e^\top \mathbf{K}_e(E_e)\mathbf{L}_e$, where
$\mathbf{L}_e$ maps global to local DOFs.

Each element carries a scalar design density $\bar{\rho}_e\in[0,1]$.
The solid isotropic material with penalization (SIMP) law assigns an effective Young's modulus
\begin{equation}
  E_e(\bar{\rho}_e)
  = E_{\min} + \bar{\rho}_e^{\,p}\,\bigl(E_0 - E_{\min}\bigr),
  \qquad
  E_{\min} = \varepsilon\, E_0,
\end{equation}
with reference modulus $E_0$, penalty exponent $p>1$, and a small positive stiffness floor
$\varepsilon\ll 1$ so that void-like regions retain a well-posed, strictly positive-definite
$\mathbf{K}$ on the free subspace.

Static equilibrium with a prescribed nodal force vector $\mathbf{f}\in\mathbb{R}^{2|\mathcal{N}|}$
and homogeneous Dirichlet data on a specified subset of displacement components (fixed DOFs)
yields the reduced linear system
\begin{equation}
  \mathbf{K}_{ff}\,\mathbf{u}_f = \mathbf{f}_f,
\end{equation}
where subscripts $f$ denote rows and columns associated with unconstrained DOFs after
partitioning.
The compliance of the structure for the given load is the elastic strain energy
\begin{equation}
  C \;=\; \mathbf{f}^\top \mathbf{u}
  \;=\; \mathbf{u}_f^\top \mathbf{K}_{ff}\,\mathbf{u}_f
  \;=\; \int_{\Omega} \boldsymbol{\sigma}^\top \boldsymbol{\varepsilon}\,\mathrm{d}A,
\end{equation}
evaluated consistently with the discrete Q4 approximation above.

\paragraph{Fluid Mechanics.} We consider a two-dimensional sectional wind tunnel on a rectangle
$\Omega=(0,L_x)\times(0,H)$ with coordinates $(x,y)$, where $y=0$ is the floor and $y=H$
the ceiling. A parameterized car silhouette is described by a single smooth graph
$y=y_{\mathrm{body}}(x)$ for $x\in(0,L_x)$. In the implementation, the upper boundary is
the sum of a fixed deck height (proportional to $H$), a Gaussian bump in $x$ controlled by
amplitude and width parameters, an algebraic rear taper proportional to
$\bigl(1-x/L_x\bigr)_+^2$, followed by a hard cap so that $y_{\mathrm{body}}\le c\,H$ with
$c<1$. The solid region used by the immersed-body mask lies below this surface,
$\{(x,y):0<y<y_{\mathrm{body}}(x)\}$, smoothed across a thin transition layer. The velocity $\mathbf{u}=(u,v)^\top$, pressure $p$, and constant density $\rho$ satisfy a
Brinkman--Stokes system posed on the entire tunnel rectangle,
\begin{equation*}
  -\mu\,\nabla^2 \mathbf{u} + \nabla p + \eta\,\chi\,\mathbf{u} = \mathbf{0},
  \qquad
  \nabla\cdot \mathbf{u} = 0,
\end{equation*}
where $\mu$ is the dynamic viscosity and $\chi(\mathbf{x})\in[0,1]$ is a smoothed indicator
with $\chi\approx 1$ in the car region and $\chi\approx 0$ in the exterior fluid, built
from a logistic mollification of $y_{\mathrm{body}}(x)-y$ with width $\tau>0$. The
penalization coefficient $\eta$ is taken mesh-dependent, scaling like
$\mu/\Delta x^2$ for a cell size $\Delta x$ comparable to $L_x/n_x$, so that explicit
pseudo-time stepping remains stable while enforcing $\mathbf{u}\approx\mathbf{0}$ where
$\chi\approx 1$. Where $\chi\approx 0$, the equations reduce to incompressible Stokes in
the exterior fluid.

A uniform streamwise inlet $u=U_\infty$, $v=0$ is imposed at $x=0$, no-slip
$\mathbf{u}=\mathbf{0}$ on the top and bottom walls, and an approximate outflow at
$x=L_x$ by extrapolating $(u,v)$ from the penultimate column (zero normal gradient of
velocity). In the discrete implementation, the no-slip row at $y=0$ is applied after the
inlet column, so the bottom-left corner is no-slip rather than plug-flow; this affects only
a single grid point at the resolution used here. Because the model is marched explicitly in a pseudo-time $\tau_{\mathrm{ac}}$, an
artificial compressibility pressure update is used,
\begin{equation*}
  \frac{\partial p}{\partial \tau_{\mathrm{ac}}} + \beta^2\,\nabla\cdot \mathbf{u} = 0,
\end{equation*}
with $\beta$ chosen proportional to $U_\infty$ in the code. The velocity is advanced with
an explicit Euler step combining viscous diffusion, pressure gradients scaled by
$\rho^{-1}$, and the Brinkman sink $-(\eta/\rho)\chi\mathbf{u}$; the pressure is then
projected to zero mean. Only a finite number of pseudo-time iterations is taken, so
the reported fields approximate a steady discrete Brinkman--Stokes--AC fixed point at
the chosen resolution rather than a fully converged continuum limit.

A streamwise drag force per unit span is estimated from the relaxed discrete fields by a
two-term volume integral that contracts $p$ and a single velocity gradient component with
gradients of $\chi$,
\begin{equation*}
  D \;\approx\;
  \rho\!\int_\Omega p\,\frac{\partial \chi}{\partial x}\,\mathrm{d}A
  \;-\;
  \mu\!\int_\Omega \frac{\partial u}{\partial y}\,\frac{\partial \chi}{\partial y}\,\mathrm{d}A,
\end{equation*}
with an overall sign convention fixed so that $D>0$ corresponds to the resistive streamwise
force used in optimization. The pressure contribution is in the spirit of Peskin-type IBM
force density $-\nabla\chi$ contracted with the pressure; the second term is a
reduced viscous proxy (it retains only $\partial u/\partial y$ rather than the full
symmetric viscous traction on the interface) and should be read as an engineering
surrogate at low mesh Reynolds number, not a high-fidelity boundary traction integral.

The surface $y_{\mathrm{body}}(x)$ is controlled by a latent vector
$\mathbf{z}\in\mathbb{R}^4$ mapped through bounded nonlinearities to a small set of shape
scales (bump amplitude, longitudinal width, streamwise center, and a signed rear taper).
This enforces a simple single-hump family suitable for few-parameter inverse design.

\subsection{Atomistic Simulations}

\paragraph{Density Functional Theory.} Kohn-Sham Density Functional Theory (KS-DFT) is a foundational computational framework that simplifies the complex, many-body interactions of electrons by mapping them onto a system of non-interacting particles, allowing researchers to predict the properties of materials based on their overall electron density rather than tracking every individual electron. We implement a radial restricted ``helium-like'' Kohn-Sham DFT simulator, i.e., a spherical closed-shell density $\rho(\mathbf{r})=\rho(r)$ on $r\in[0,R]$ with a single nuclear center. The reduced radial amplitude $u(r)$ (with $u(0)=u(R)=0$) defines a restricted two-electron mean field via $\rho(r)=\frac{N_e/2}{2\pi r^2}u(r)^2$ and $\int 4\pi r^2\rho\,\mathrm{d}r=N_e$ (intended for $N_e{=}2$, i.e.\ one doubly occupied $s$-like spatial orbital). The non-interacting kinetic energy uses $-\tfrac12\,\mathrm{d}^2/\mathrm{d}r^2$ on $u$ (uniform radial mesh in the implementation). The external potential is a softened Coulomb $V_{\mathrm{ext}}=-Z/\sqrt{r^2+\delta^2}$; $V_H$ solves the spherical Poisson equation for $\rho$; $V_{\mathrm{xc}}$ comes from a local $\varepsilon_{\mathrm{xc}}(\rho)$ (Slater exchange plus an analytic correlation surrogate). For Kohn-Sham self-consistency, $V_{\mathrm{eff}}=V_{\mathrm{ext}}+V_H+V_{\mathrm{xc}}$ and iterating between $\rho$ and $u$ with linear mixing are used. The total energy is $E=T_s+E_{\mathrm{ext}}+\tfrac12 E_H+E_{\mathrm{xc}}$ with the same $4\pi r^2\,\mathrm{d}r$ measure.

\paragraph{Tight Binding.} The Tight-Binding (TB) approximation replaces the full single-particle Hamiltonian by a sparse matrix in a minimal local basis: one orthogonal orbital per site, nearest-neighbor hopping only, and no explicit spin degrees of freedom unless stated otherwise. We consider a single spinless orbital on each of $N\ge 3$ sites arranged on a ring (one spatial dimension, nearest neighbors only). Onsite energies $\varepsilon_n$ and real hopping magnitudes $t_{n,n+1}>0$ enter the Hermitian Hamiltonian with matrix elements
\begin{equation}
  H_{nn}=\varepsilon_n,\qquad
  H_{n,n+1}=H_{n+1,n}=-\,t_{n,n+1}\qquad (n=0,\ldots,N-2),
\end{equation}
and a closing bond between sites $N-1$ and $0$ with the same convention, $H_{N-1,0}=H_{0,N-1}=-\,t_{N-1,0}$ in the purely periodic problem. Real $t$ and this sign choice correspond to zero magnetic flux through the ring in the usual tight-binding sense (no Peierls phases on the intracell bonds).

To embed this $N$-site cluster as one supercell of an infinite 1D periodic chain, let $a$ denote the nearest-neighbor distance along the backbone and $A=N a$ the supercell length. A crystal wavevector $k$ enters only through the intercell link that reconnects site $N{-}1$ to site $0$ of the next cell. The Bloch-reduced $N\times N$ Hamiltonian $H(k)$ keeps the same intracell matrix elements as above, while the wrap becomes
\begin{equation}
  H_{N-1,0}(k)=-\,t_{N-1,0}\,e^{ikA},\qquad
  H_{0,N-1}(k)=-\,t_{N-1,0}\,e^{-ikA},
\end{equation}
so $H(k)$ is Hermitian for real $k$, and $H(k+2\pi/A)=H(k)$. For each $k$, the single-particle energies are the eigenvalues of $H(k)$; they may be sorted ascending at fixed $k$ for plotting. The finite supercell spectrum used for total energies and eigenvectors at the supercell origin is $H(0)$, i.e.\ the same ring Hamiltonian without a phase on the closing bond. The total density of states per site is the spectral measure of $H(0)$,
\begin{equation}
  D(E)=\frac{1}{N}\sum_{j=1}^{N}\delta\bigl(E-E_j\bigr),
\end{equation}
with $\{E_j\}$ the eigenvalues of $H(0)$. In the benchmarked parameter-identification task, the loss matches $\{E_j\}$ and the Bloch band matrix $E(k)$ to a frozen reference (weights $1$ and $0.25$); a Gaussian-smoothed DOS is available from the same spectrum for plotting only.

\subsection{Rendering}

\paragraph{Gaussian Splatting.}
We consider a fixed image grid of size $H \times W$ over a normalized square domain $\mathcal{D} = [-1,1]^2$. Each pixel center is denoted $\mathbf{u}_{ij} \in \mathcal{D}$ for $i \in \{1,\ldots,H\}$, $j \in \{1,\ldots,W\}$. The renderer places $K$ splats. Splat $k$ is specified by a center $\mathbf{x}_k \in \mathbb{R}^2$, a positive scale $\sigma_k$, an opacity $\alpha_k \in (0,1)$, and a diffuse color $\mathbf{c}_k \in (0,1)^3$. Let $r_k^2(\mathbf{u}) = \|\mathbf{u} - \mathbf{x}_k\|^2$. The splat weight on the image plane is
\begin{equation}
  g_k(\mathbf{u}) = \exp\!\left(-\,\frac{r_k^2(\mathbf{u})}{2\sigma_k^2}\right),
  \qquad
  w_k(\mathbf{u}) = \alpha_k\, g_k(\mathbf{u}).
\end{equation}
Thus $w_k$ is non-negative and peaks at $\mathbf{x}_k$ with characteristic width $\sigma_k$. In the implementation, $\sigma_k$ is obtained from a unconstrained parameter via a smooth positive map (softplus) plus a small floor, so that $\sigma_k$ stays strictly positive and gradients remain stable.

At each $\mathbf{u}$, splat colors are combined with a normalized weighted average of the diffuse colors:
\begin{equation}
  \widehat{\mathbf{c}}(\mathbf{u}) =
  \frac{\sum_{k=1}^{K} w_k(\mathbf{u})\, \mathbf{c}_k}
       {\sum_{k=1}^{K} w_k(\mathbf{u}) + \varepsilon},
\end{equation}
with a fixed small $\varepsilon > 0$ only to avoid division by zero in regions where all weights vanish. When $\sum_k w_k(\mathbf{u}) > 0$, $\widehat{\mathbf{c}}(\mathbf{u})$ lies in the convex hull of the $\mathbf{c}_k$ because the $w_k$ are non-negative. Splats are blended in an orderless way: no explicit back-to-front sort is required. Let $S(\mathbf{u}) = \sum_{k=1}^{K} w_k(\mathbf{u})$ act as a non-negative accumulated opacity density at $\mathbf{u}$. A smooth coverage is
\begin{equation}
  A(\mathbf{u}) = 1 - \mathrm{e}^{-S(\mathbf{u})} \in [0,1).
\end{equation}
For large $S$, $A \to 1$; for $S \to 0$, $A \sim S$. Given a constant background color $\mathbf{I}_{\mathrm{bg}}$ (scalar gray or RGB), the displayed image is
\begin{equation}
  \mathbf{I}(\mathbf{u}) =
  A(\mathbf{u})\, \widehat{\mathbf{c}}(\mathbf{u})
  + \bigl(1 - A(\mathbf{u})\bigr)\, \mathbf{I}_{\mathrm{bg}}.
\end{equation}

The continuous fields $g_k$, $w_k$, $S$, $A$, $\widehat{\mathbf{c}}$, and
$\mathbf{I}$ are evaluated on the regular grid $\{\mathbf{u}_{ij}\}$ only. Coordinates $\mathbf{u}_{ij}$ are generated by uniform sampling of $[-1,1]$ along each axis (tensor grid), so pixel spacing is uniform in the normalized frame. All operations (Gaussian evaluation, sums, exponentials, and the final blend) are applied pointwise on that grid; the result is an $H \times W \times 3$ RGB image in $[0,1]^3$ suitable as a differentiable forward map in autodiff frameworks.

\paragraph{Atmospheric Scattering.}
Sunlight entering Earth's atmosphere interacts with air molecules and aerosols through scattering and absorption Scattering redirects photons out of the incident beam; absorption removes energy from the beam and heats the medium. Because molecular (Rayleigh) scattering scales strongly with wavelength, short wavelengths are scattered more efficiently than long ones, which is why clear skies appear blue and sunsets appear reddened when long optical paths sample more scattering and extinction. Larger particles (haze, cloud droplets) contribute Mie scattering, which is more forward-peaked and less spectrally selective, producing whitish hazes and bright glints around the sun. For rendering and inverse problems, the full radiative transfer equation couples emission, absorption, and multiple scattering; practical sky models often approximate the visible sky by single scattering along a view ray plus analytic or tabulated optical depths, trading physical completeness for speed and differentiability while preserving the main perceptual cues (gradient, polarization of color with elevation, solar disk glow).

The scene is a concentric-shell atmosphere between planet radius $R_g$ and outer radius $R_t$ in a planet-centered frame. Pinhole rays $(\mathbf{o},\hat{\mathbf{d}})$ are intersected with the outer sphere; integration runs from entry to exit along $\hat{\mathbf{d}}$, truncated at the first intersection with the ground sphere of radius $R_g$ so samples stay above the surface. At a point $\mathbf{p}$ with altitude $h=\max(\|\mathbf{p}\|-R_g,0)$, scalar number densities follow exponential atmospheres $\rho_R(h)=\exp(-h/H_R)$ and $\rho_M(h)=\exp(-h/H_M)$, with scale heights $H_R,H_M$.
Per-channel extinction uses RGB coefficients $\boldsymbol{\beta}_R,\boldsymbol{\beta}_M$ (1/m) and an optional absorption vector $\boldsymbol{\beta}_A$ weighted by the same $\rho_R$ factor:
\begin{equation}
\boldsymbol{\sigma}_t(\mathbf{p})=\rho_R(h)\,\boldsymbol{\beta}_R+\rho_M(h)\,\boldsymbol{\beta}_M+\rho_R(h)\,\boldsymbol{\beta}_A.
\end{equation}
Along the view ray, samples are spaced uniformly in distance; a cumulative sum approximates
$\boldsymbol{\tau}_{\mathrm{view}}(s)=\int_{s_{\min}}^{s}\boldsymbol{\sigma}_t(\mathbf{o}+s'\hat{\mathbf{d}})\,\mathrm{d}s'$.
From each sample $\mathbf{p}$, a second quadrature marches toward the (unit) sun direction $\hat{\mathbf{s}}$ out to the atmosphere shell, yielding $\boldsymbol{\tau}_{\odot}(\mathbf{p})$ for sunlight attenuation to~$\mathbf{p}$. Let $\mu=\hat{\mathbf{d}}\!\cdot\!\hat{\mathbf{s}}$.
Rayleigh uses $P_R(\mu)=\frac{3}{16\pi}(1+\mu^2)$; Mie uses the Henyey--Greenstein phase $P_M(\mu\mid g)$.
The in-scattered radiance accumulated along the camera ray is a discrete sum of
\begin{equation}
\Big(\rho_R\,\boldsymbol{\beta}_R P_R(\mu)+\rho_M\,\boldsymbol{\beta}_M P_M(\mu\mid g)\Big)\,
\exp\!\big(-(\boldsymbol{\tau}_{\mathrm{view}}+\boldsymbol{\tau}_{\odot})\big)\,\Delta s,
\end{equation}
scaled by a per-channel sun intensity $\mathbf{I}_\odot$, exposure, optional white-balance normalization, and mapped to $[0,1]$ by $\mathbf{L}\mapsto 1-\exp(-\mathbf{L})$ (invalid rays return black).

\subsection{Semi-Empirical Physics Models}

\paragraph{Tire Magic Formula.}
The magic formula is a semi-empirical model for tire dynamics which describes the relationship between the slip ratio and slip angle of a tire and the resulting longitudinal and lateral forces. The model is widely used in vehicle dynamics simulations and control systems. The core equation of the magic formula can be expressed as follows:
\begin{align}
y = D \sin(C \arctan(B x - E (B x - \arctan(B x)))).
\end{align}
It is used to model the longitudinal force as a function of the slip ratio, in which case $x$ is the slip ratio $\kappa$, and $y$ is the longitudinal force $F_x$. It is also used to model the lateral force as a function of the slip angle, in which case $x$ is the slip angle $\alpha$, and $y$ is the lateral force $F_y$. The parameters $B$, $C$, $D$, and $E$ can be tuned to fit experimental data, allowing the magic formula to accurately capture the complex behavior of tire forces under various conditions.

\paragraph{Battery Thevenin Model.}
The Thevenin model simplifies the complex electrochemical processes in a battery into an equivalent circuit consisting of a voltage source and resistors. We use a first-order Thevenin model with one RC polarization branch. Let $I(t)$ denote the applied current at time $t$ (positive for discharge in the usual sign convention; the benchmark uses bipolar random pulse profiles). Let $Q$ denote the nominal capacity and $\eta\in(0,1]$ denote the coulombic efficiency. Let $R_0$ denote ohmic resistance, and $(R_1,C_1)$ characterize polarization dynamics. The model state is the state of charge $\mathrm{SOC}(t)$ and the RC polarization voltage $V_{\mathrm{rc}}(t)$:
\begin{align}
\dot{\mathrm{SOC}}(t) &= -\,\eta\,\frac{I(t)}{Q},\\
\dot V_{\mathrm{rc}}(t) &= -\frac{1}{R_1C_1}V_{\mathrm{rc}}(t) + \frac{1}{C_1}I(t),
\end{align}
and the terminal voltage is
\begin{equation}
V(t)=V_{\mathrm{oc}}(\mathrm{SOC}(t)) - R_0 I(t) - V_{\mathrm{rc}}(t).
\end{equation}
The open-circuit voltage is parameterized as a smooth SoC map; in our implementation we use a cubic polynomial
\begin{equation}
V_{\mathrm{oc}}(s)=a_0+a_1 s+a_2 s^2+a_3 s^3,\qquad s\in[0,1].
\end{equation}
For simulation, this continuous-time model is discretized at step size $\Delta t$ and integrated in a differentiable manner, with SoC constrained to $[0,1]$.
In the parameter-identification task, both the circuit tuple $(Q,\eta,R_0,R_1,C_1)$ and the OCV coefficients $(a_0,\ldots,a_3)$ are learnable; the objective is $\mathrm{MSE}(V)+0.2\,\mathrm{MSE}(\mathrm{SOC})$ against trajectories from a frozen ground-truth model driven by the same currents and initial SoC.


\section{Details of Experiment Setups and Results}
\label{app:experiment}

\subsection{Experiment Setups}

For each task and each of the 12 optimizers, we evaluate a shared hyper-parameter grid (207 configurations in total across optimizers) and select the best configuration by minimizing the mean best loss over random initializations. Each configuration is repeated 3 times, and each run executes $T=100$ optimizer epochs. One epoch is one call to the optimizer's update rule, with the scalar loss defined by the task's loss function. Experiments are run on an NVIDIA A100 GPU (80\,GB).
For the SIMP topology task only, a run is treated as non-converged if its best loss is negative (physically inconsistent with the non-negative compliance-plus-volume objective). Non-converged repeats are excluded from hyper-parameter selection and from the reported mean/std; selection otherwise follows the same minimize-mean-best-loss rule on the remaining repeats.

The benchmarked tasks match the registry defaults used in the grid search. In particular: battery parameter identification minimizes $\mathrm{MSE}(V)+0.2\,\mathrm{MSE}(\mathrm{SOC})$ under fixed pulse currents; tight-binding fitting minimizes $\mathrm{MSE}(\{E_j\})+0.25\,\mathrm{MSE}(E(k))$ on supercell eigenvalues and Bloch bands; projectile inverse design minimizes $(s_x-16.974)^2$ over the launch angle; SIMP topology design uses $\mathcal{L}=C/C_{\mathrm{ref}}+25\,(\overline{\rho}-0.5)^2$ on a $30\times20$ half MBB beam; flight open-loop control maximizes downrange via $\mathcal{L}=-x_T/U+(2\times10^{-5}/U)\,\mathrm{mean}(\mathbf{u}^{\odot 2})+(1\times10^{-5}/U)\,\mathrm{mean}((\mathbf{u}_{t+1}-\mathbf{u}_t)^{\odot 2})$ with $U=10^{6}\,\mathrm{m}$; rigid-body control optimizes a $(10\times7)$-dimensional open-loop actuation table toward goal $(0.9,0.15)$; and mass--spring control optimizes a $(100\times11)$-dimensional table toward $(0.9,0.2)$.

Let $\theta^{(0)}$ denote the baseline design-variable tensors, i.e., floating-point entries of \texttt{model.named\_parameters()} only. Simulator buffers such as FEM stiffness stacks, nodal forces, and mesh coordinates are copied unchanged. For run $r$, each design variable $\theta$ is reinitialized as
\begin{equation}
  \tilde{\theta} = \theta + \sigma_{\mathrm{init}}\cdot \max\bigl(m(\theta),\,10^{-2}\bigr)\cdot \xi,
  \qquad \xi \sim \mathcal{N}(0,I),
\end{equation}
where $m(\theta)$ is the mean of $|\theta|$, and $\sigma_{\mathrm{init}}$ is the init noise (set to 0.05). Noise samples use a \texttt{torch.Generator} with seed 0.

Table~\ref{tab:bench-opt-specs} lists the searched hyper-parameters and fixed secondary knobs for each optimizer. Candidate values are drawn from shared global grids: $\texttt{lr}/\texttt{polish\_lr}\in\{10^{-3},10^{-2},10^{-1},1\}$, $\texttt{momentum}/\texttt{beta1}\in\{0,0.9\}$, $\texttt{damping}\in\{10^{-4},10^{-3},10^{-2}\}$, $\texttt{sigma}/\texttt{init\_sigma}\in\{0.05,0.1,0.5\}$, $\texttt{population}\in\{8,16,32\}$, $\texttt{temperature}\in\{0.5,1,2\}$. For approximate second-order methods, each outer step uses the autograd gradient from the task's forward--backward pass. For global optimizers, after each global search proposal, the optimizers optionally run a short gradient-descent polish (with \texttt{polish\_steps=5} fixed) to locally refine the best candidate.

\begin{table}[t]
  \centering
  \caption{Optimizer hyper-parameter search space for the benchmark.}
  \label{tab:bench-opt-specs}
  \begin{tabular}{@{}lp{0.62\linewidth}@{}}
    \toprule
    Name & Searched / fixed configurations \\
    \midrule
    SGD & search \texttt{lr}, \texttt{momentum} (8 configs) \\
    RMSprop & search \texttt{lr}, \texttt{momentum} (8 configs) \\
    Adam & search \texttt{lr}, \texttt{beta1}; fixed \texttt{beta2=0.999} (8 configs) \\
    AdamW & search \texttt{lr}, \texttt{beta1}; fixed \texttt{beta2=0.999} (8 configs) \\
    LBFGS & search \texttt{lr}; fixed \texttt{max\_iter=10}, \texttt{line\_search\_fn=strong\_wolfe} (4 configs) \\
    KFAC & search \texttt{lr}, \texttt{damping}; fixed \texttt{momentum=0.9}, \texttt{beta=0.95}, \texttt{precondition\_frequency=10} (12 configs) \\
    Shampoo & search \texttt{lr}, \texttt{damping}; fixed \texttt{beta=0.95}, \texttt{precondition\_frequency=10} (12 configs) \\
    SOAP & search \texttt{lr}, \texttt{damping}; fixed \texttt{betas=(0.9,0.99)}, \texttt{beta\_precond=0.95}, \texttt{precondition\_frequency=10} (12 configs) \\
    RandomSearch & search \texttt{sigma}, \texttt{candidates\_per\_step}, \texttt{polish\_lr}; fixed \texttt{polish\_steps=5}, \texttt{clamp=10} (36 configs) \\
    SimulatedAnnealing & search \texttt{sigma}, \texttt{temperature}, \texttt{candidates\_per\_step}; fixed \texttt{cooling=0.99}, \texttt{polish\_steps=5}, \texttt{polish\_lr=1e-2}, \texttt{clamp=10} (27 configs) \\
    ParticleSwarm & search \texttt{init\_sigma}, \texttt{swarm\_size}, \texttt{polish\_lr}; fixed \texttt{polish\_steps=5}, \texttt{clamp=10} (36 configs) \\
    DifferentialEvolution & search \texttt{init\_sigma}, \texttt{population\_size}, \texttt{polish\_lr}; fixed \texttt{polish\_steps=5}, \texttt{clamp=10} (36 configs) \\
    \bottomrule
  \end{tabular}
\end{table}

Exceptions inside the inner training loop are caught per run: the failure is logged, and the benchmark continues with the next repeat or optimizer without terminating the entire sweep.

\subsection{Compute Resources}
\label{app:compute_resources}

We record the wall time and peak GPU memory usage for each run of the selected (best) hyper-parameter configuration. The wall time is measured for the full optimization run and converted to a per-step average by dividing by the number of effective epochs; mean/std are computed over the three repeats. Peak GPU memory is likewise averaged over those three best-configuration repeats (reported in MiB as ``MB'' in the tables).

\begin{table}[tbh]
\centering
\caption{Compute Resource Usage for DFT XC Functional Parameter Fitting.}
\begin{tabular}{@{}lcc@{}}
\toprule
Optimizer & Step wall (s), mean $\pm$ std & Peak GPU (MB), mean $\pm$ std \\
\midrule
SGD & $1.1 \times 10^{0} \pm 3.1 \times 10^{-2}$ & $5.1 \times 10^{1} \pm 0.0$ \\
RMSprop & $1.1 \times 10^{0} \pm 8.3 \times 10^{-3}$ & $5.1 \times 10^{1} \pm 0.0$ \\
Adam & $9.6 \times 10^{-1} \pm 1.1 \times 10^{-2}$ & $5.1 \times 10^{1} \pm 0.0$ \\
AdamW & $9.6 \times 10^{-1} \pm 1.9 \times 10^{-2}$ & $5.1 \times 10^{1} \pm 0.0$ \\
LBFGS & $6.4 \times 10^{-1} \pm 1.6 \times 10^{-2}$ & $5.1 \times 10^{1} \pm 1.8 \times 10^{-3}$ \\
KFAC & $1.1 \times 10^{0} \pm 1.2 \times 10^{-2}$ & $5.1 \times 10^{1} \pm 0.0$ \\
Shampoo & $1.1 \times 10^{0} \pm 3.3 \times 10^{-3}$ & $5.1 \times 10^{1} \pm 0.0$ \\
SOAP & $1.1 \times 10^{0} \pm 5.0 \times 10^{-3}$ & $5.1 \times 10^{1} \pm 0.0$ \\
RandomSearch & $1.9 \times 10^{1} \pm 1.6 \times 10^{-1}$ & $5.1 \times 10^{1} \pm 0.0$ \\
SimulatedAnnealing & $1.1 \times 10^{1} \pm 3.7 \times 10^{-2}$ & $5.1 \times 10^{1} \pm 0.0$ \\
ParticleSwarm & $5.0 \times 10^{0} \pm 1.4 \times 10^{-2}$ & $5.1 \times 10^{1} \pm 0.0$ \\
DifferentialEvolution & $5.0 \times 10^{0} \pm 1.8 \times 10^{-2}$ & $5.1 \times 10^{1} \pm 0.0$ \\
\bottomrule
\end{tabular}
\end{table}

\begin{table}[tbh]
\centering
\caption{Compute Resource Usage for Tire Magic Formula Parameter Fitting.}
\begin{tabular}{@{}lcc@{}}
\toprule
Optimizer & Step wall (s), mean $\pm$ std & Peak GPU (MB), mean $\pm$ std \\
\midrule
SGD & $2.2 \times 10^{-3} \pm 8.2 \times 10^{-4}$ & $8.7 \times 10^{0} \pm 0.0$ \\
RMSprop & $2.0 \times 10^{-3} \pm 5.0 \times 10^{-5}$ & $8.7 \times 10^{0} \pm 0.0$ \\
Adam & $2.4 \times 10^{-3} \pm 1.7 \times 10^{-4}$ & $8.7 \times 10^{0} \pm 0.0$ \\
AdamW & $2.6 \times 10^{-3} \pm 2.3 \times 10^{-4}$ & $8.7 \times 10^{0} \pm 0.0$ \\
LBFGS & $6.4 \times 10^{-1} \pm 8.4 \times 10^{-2}$ & $8.9 \times 10^{0} \pm 9.2 \times 10^{-4}$ \\
KFAC & $2.8 \times 10^{-3} \pm 2.0 \times 10^{-4}$ & $8.7 \times 10^{0} \pm 0.0$ \\
Shampoo & $2.6 \times 10^{-3} \pm 4.7 \times 10^{-5}$ & $8.7 \times 10^{0} \pm 0.0$ \\
SOAP & $2.6 \times 10^{-3} \pm 9.7 \times 10^{-5}$ & $8.7 \times 10^{0} \pm 0.0$ \\
RandomSearch & $7.8 \times 10^{-2} \pm 1.3 \times 10^{-3}$ & $8.7 \times 10^{0} \pm 0.0$ \\
SimulatedAnnealing & $1.2 \times 10^{-1} \pm 6.5 \times 10^{-4}$ & $8.7 \times 10^{0} \pm 0.0$ \\
ParticleSwarm & $1.0 \times 10^{-1} \pm 2.9 \times 10^{-3}$ & $8.7 \times 10^{0} \pm 0.0$ \\
DifferentialEvolution & $1.8 \times 10^{-1} \pm 1.1 \times 10^{-3}$ & $8.7 \times 10^{0} \pm 0.0$ \\
\bottomrule
\end{tabular}
\end{table}

\begin{table}[tbh]
\centering
\caption{Compute Resource Usage for Battery Thevenin Model Parameter Fitting.}
\begin{tabular}{@{}lcc@{}}
\toprule
Optimizer & Step wall (s), mean $\pm$ std & Peak GPU (MB), mean $\pm$ std \\
\midrule
SGD & $9.1 \times 10^{-2} \pm 4.0 \times 10^{-4}$ & $9.1 \times 10^{0} \pm 0.0$ \\
RMSprop & $8.6 \times 10^{-2} \pm 1.4 \times 10^{-3}$ & $9.1 \times 10^{0} \pm 0.0$ \\
Adam & $8.6 \times 10^{-2} \pm 8.1 \times 10^{-4}$ & $9.1 \times 10^{0} \pm 0.0$ \\
AdamW & $8.7 \times 10^{-2} \pm 7.8 \times 10^{-4}$ & $9.1 \times 10^{0} \pm 0.0$ \\
LBFGS & $6.4 \times 10^{-1} \pm 2.4 \times 10^{-2}$ & $9.2 \times 10^{0} \pm 6.3 \times 10^{-3}$ \\
KFAC & $9.3 \times 10^{-2} \pm 1.7 \times 10^{-3}$ & $9.1 \times 10^{0} \pm 0.0$ \\
Shampoo & $8.8 \times 10^{-2} \pm 4.9 \times 10^{-4}$ & $9.1 \times 10^{0} \pm 0.0$ \\
SOAP & $8.7 \times 10^{-2} \pm 1.2 \times 10^{-3}$ & $9.1 \times 10^{0} \pm 0.0$ \\
RandomSearch & $8.1 \times 10^{-1} \pm 2.0 \times 10^{-2}$ & $9.1 \times 10^{0} \pm 0.0$ \\
SimulatedAnnealing & $8.7 \times 10^{-1} \pm 1.7 \times 10^{-2}$ & $9.1 \times 10^{0} \pm 0.0$ \\
ParticleSwarm & $8.0 \times 10^{-1} \pm 1.5 \times 10^{-2}$ & $9.1 \times 10^{0} \pm 0.0$ \\
DifferentialEvolution & $1.5 \times 10^{0} \pm 6.6 \times 10^{-3}$ & $9.1 \times 10^{0} \pm 0.0$ \\
\bottomrule
\end{tabular}
\end{table}

\begin{table}[tbh]
\centering
\caption{Compute Resource Usage for Tight Binding Parameter Fitting.}
\begin{tabular}{@{}lcc@{}}
\toprule
Optimizer & Step wall (s), mean $\pm$ std & Peak GPU (MB), mean $\pm$ std \\
\midrule
SGD & $6.8 \times 10^{-1} \pm 1.6 \times 10^{-3}$ & $1.8 \times 10^{1} \pm 0.0$ \\
RMSprop & $6.9 \times 10^{-1} \pm 9.0 \times 10^{-3}$ & $1.8 \times 10^{1} \pm 0.0$ \\
Adam & $8.1 \times 10^{-1} \pm 9.8 \times 10^{-2}$ & $1.8 \times 10^{1} \pm 0.0$ \\
AdamW & $1.2 \times 10^{0} \pm 5.5 \times 10^{-3}$ & $1.8 \times 10^{1} \pm 0.0$ \\
LBFGS & $2.0 \times 10^{0} \pm 2.1 \times 10^{-2}$ & $1.8 \times 10^{1} \pm 4.8 \times 10^{-3}$ \\
KFAC & $1.1 \times 10^{0} \pm 5.7 \times 10^{-3}$ & $1.8 \times 10^{1} \pm 0.0$ \\
Shampoo & $1.2 \times 10^{0} \pm 7.1 \times 10^{-3}$ & $1.8 \times 10^{1} \pm 0.0$ \\
SOAP & $1.2 \times 10^{0} \pm 5.8 \times 10^{-3}$ & $1.8 \times 10^{1} \pm 0.0$ \\
RandomSearch & $1.1 \times 10^{1} \pm 2.8 \times 10^{-1}$ & $1.8 \times 10^{1} \pm 0.0$ \\
SimulatedAnnealing & $5.8 \times 10^{0} \pm 6.5 \times 10^{-3}$ & $1.8 \times 10^{1} \pm 0.0$ \\
ParticleSwarm & $1.2 \times 10^{1} \pm 3.5 \times 10^{-2}$ & $1.8 \times 10^{1} \pm 0.0$ \\
DifferentialEvolution & $5.6 \times 10^{0} \pm 1.4 \times 10^{-2}$ & $1.8 \times 10^{1} \pm 0.0$ \\
\bottomrule
\end{tabular}
\end{table}

\begin{table}[tbh]
\centering
\caption{Compute Resource Usage for Gaussian Splatting 2D Reconstruction.}
\begin{tabular}{@{}lcc@{}}
\toprule
Optimizer & Step wall (s), mean $\pm$ std & Peak GPU (MB), mean $\pm$ std \\
\midrule
SGD & $6.8 \times 10^{-3} \pm 1.1 \times 10^{-4}$ & $1.2 \times 10^{1} \pm 0.0$ \\
RMSprop & $6.8 \times 10^{-3} \pm 7.0 \times 10^{-5}$ & $1.2 \times 10^{1} \pm 0.0$ \\
Adam & $7.2 \times 10^{-3} \pm 1.7 \times 10^{-4}$ & $1.2 \times 10^{1} \pm 0.0$ \\
AdamW & $7.1 \times 10^{-3} \pm 8.9 \times 10^{-5}$ & $1.2 \times 10^{1} \pm 0.0$ \\
LBFGS & $3.2 \times 10^{0} \pm 1.6 \times 10^{-1}$ & $1.2 \times 10^{1} \pm 0.0$ \\
KFAC & $2.5 \times 10^{-2} \pm 4.1 \times 10^{-4}$ & $1.2 \times 10^{1} \pm 0.0$ \\
Shampoo & $1.4 \times 10^{-2} \pm 1.1 \times 10^{-4}$ & $1.2 \times 10^{1} \pm 0.0$ \\
SOAP & $1.5 \times 10^{-2} \pm 4.0 \times 10^{-4}$ & $1.2 \times 10^{1} \pm 0.0$ \\
RandomSearch & $1.6 \times 10^{-1} \pm 2.2 \times 10^{-3}$ & $1.2 \times 10^{1} \pm 0.0$ \\
SimulatedAnnealing & $1.7 \times 10^{-1} \pm 6.6 \times 10^{-4}$ & $1.2 \times 10^{1} \pm 0.0$ \\
ParticleSwarm & $3.9 \times 10^{-1} \pm 9.8 \times 10^{-3}$ & $1.2 \times 10^{1} \pm 0.0$ \\
DifferentialEvolution & $2.2 \times 10^{-1} \pm 7.7 \times 10^{-4}$ & $1.2 \times 10^{1} \pm 0.0$ \\
\bottomrule
\end{tabular}
\end{table}

\begin{table}[tbh]
\centering
\caption{Compute Resource Usage for Projectile Motion Range Matching.}
\begin{tabular}{@{}lcc@{}}
\toprule
Optimizer & Step wall (s), mean $\pm$ std & Peak GPU (MB), mean $\pm$ std \\
\midrule
SGD & $1.0 \times 10^{-3} \pm 8.3 \times 10^{-6}$ & $8.7 \times 10^{0} \pm 0.0$ \\
RMSprop & $1.3 \times 10^{-3} \pm 6.3 \times 10^{-5}$ & $8.7 \times 10^{0} \pm 0.0$ \\
Adam & $3.5 \times 10^{-3} \pm 2.7 \times 10^{-3}$ & $8.7 \times 10^{0} \pm 0.0$ \\
AdamW & $1.8 \times 10^{-3} \pm 4.0 \times 10^{-5}$ & $8.7 \times 10^{0} \pm 0.0$ \\
LBFGS & $6.4 \times 10^{-2} \pm 2.5 \times 10^{-4}$ & $8.7 \times 10^{0} \pm 0.0$ \\
KFAC & $1.8 \times 10^{-3} \pm 1.2 \times 10^{-4}$ & $8.7 \times 10^{0} \pm 0.0$ \\
Shampoo & $1.8 \times 10^{-3} \pm 6.6 \times 10^{-5}$ & $8.7 \times 10^{0} \pm 0.0$ \\
SOAP & $1.7 \times 10^{-3} \pm 1.2 \times 10^{-4}$ & $8.7 \times 10^{0} \pm 0.0$ \\
RandomSearch & $6.5 \times 10^{-2} \pm 4.6 \times 10^{-4}$ & $8.7 \times 10^{0} \pm 0.0$ \\
SimulatedAnnealing & $1.1 \times 10^{-1} \pm 2.3 \times 10^{-3}$ & $8.7 \times 10^{0} \pm 0.0$ \\
ParticleSwarm & $2.7 \times 10^{-2} \pm 4.6 \times 10^{-4}$ & $8.7 \times 10^{0} \pm 0.0$ \\
DifferentialEvolution & $1.1 \times 10^{-1} \pm 8.4 \times 10^{-4}$ & $8.7 \times 10^{0} \pm 0.0$ \\
\bottomrule
\end{tabular}
\end{table}

\begin{table}[tbh]
\centering
\caption{Compute Resource Usage for Fluid Mechanics Shape Optimization.}
\begin{tabular}{@{}lcc@{}}
\toprule
Optimizer & Step wall (s), mean $\pm$ std & Peak GPU (MB), mean $\pm$ std \\
\midrule
SGD & $4.9 \times 10^{-1} \pm 1.5 \times 10^{-2}$ & $9.3 \times 10^{0} \pm 0.0$ \\
RMSprop & $4.8 \times 10^{-1} \pm 7.0 \times 10^{-3}$ & $9.3 \times 10^{0} \pm 0.0$ \\
Adam & $5.2 \times 10^{-1} \pm 3.1 \times 10^{-3}$ & $9.3 \times 10^{0} \pm 0.0$ \\
AdamW & $5.1 \times 10^{-1} \pm 2.5 \times 10^{-3}$ & $9.3 \times 10^{0} \pm 0.0$ \\
LBFGS & $6.6 \times 10^{-1} \pm 2.8 \times 10^{-2}$ & $9.4 \times 10^{0} \pm 5.0 \times 10^{-3}$ \\
KFAC & $5.2 \times 10^{-1} \pm 1.9 \times 10^{-2}$ & $9.3 \times 10^{0} \pm 0.0$ \\
Shampoo & $4.8 \times 10^{-1} \pm 4.6 \times 10^{-3}$ & $9.3 \times 10^{0} \pm 0.0$ \\
SOAP & $5.1 \times 10^{-1} \pm 1.9 \times 10^{-3}$ & $9.3 \times 10^{0} \pm 0.0$ \\
RandomSearch & $3.6 \times 10^{0} \pm 1.0 \times 10^{-2}$ & $9.3 \times 10^{0} \pm 0.0$ \\
SimulatedAnnealing & $6.5 \times 10^{0} \pm 6.2 \times 10^{-3}$ & $9.3 \times 10^{0} \pm 0.0$ \\
ParticleSwarm & $4.5 \times 10^{0} \pm 4.1 \times 10^{-2}$ & $9.3 \times 10^{0} \pm 0.0$ \\
DifferentialEvolution & $4.8 \times 10^{0} \pm 1.8 \times 10^{-2}$ & $9.3 \times 10^{0} \pm 0.0$ \\
\bottomrule
\end{tabular}
\end{table}

\begin{table}[tbh]
\centering
\caption{Compute Resource Usage for Atmospheric Scattering Chromaticity Maximization.}
\begin{tabular}{@{}lcc@{}}
\toprule
Optimizer & Step wall (s), mean $\pm$ std & Peak GPU (MB), mean $\pm$ std \\
\midrule
SGD & $1.5 \times 10^{-2} \pm 3.4 \times 10^{-4}$ & $3.8 \times 10^{1} \pm 0.0$ \\
RMSprop & $1.4 \times 10^{-2} \pm 8.5 \times 10^{-5}$ & $3.8 \times 10^{1} \pm 0.0$ \\
Adam & $1.4 \times 10^{-2} \pm 7.3 \times 10^{-5}$ & $3.8 \times 10^{1} \pm 0.0$ \\
AdamW & $1.4 \times 10^{-2} \pm 2.2 \times 10^{-4}$ & $3.8 \times 10^{1} \pm 0.0$ \\
LBFGS & $5.0 \times 10^{-1} \pm 4.1 \times 10^{-2}$ & $3.8 \times 10^{1} \pm 4.9 \times 10^{-3}$ \\
KFAC & $3.1 \times 10^{-2} \pm 3.1 \times 10^{-4}$ & $3.8 \times 10^{1} \pm 0.0$ \\
Shampoo & $1.6 \times 10^{-2} \pm 1.4 \times 10^{-4}$ & $3.8 \times 10^{1} \pm 0.0$ \\
SOAP & $1.6 \times 10^{-2} \pm 2.0 \times 10^{-4}$ & $3.8 \times 10^{1} \pm 0.0$ \\
RandomSearch & $2.8 \times 10^{-1} \pm 2.2 \times 10^{-3}$ & $3.8 \times 10^{1} \pm 0.0$ \\
SimulatedAnnealing & $4.0 \times 10^{-1} \pm 3.5 \times 10^{-4}$ & $3.8 \times 10^{1} \pm 0.0$ \\
ParticleSwarm & $5.0 \times 10^{-1} \pm 3.3 \times 10^{-3}$ & $3.8 \times 10^{1} \pm 0.0$ \\
DifferentialEvolution & $8.9 \times 10^{-1} \pm 5.4 \times 10^{-3}$ & $3.8 \times 10^{1} \pm 0.0$ \\
\bottomrule
\end{tabular}
\end{table}

\begin{table}[tbh]
\centering
\caption{Compute Resource Usage for Elastic Mechanics Topology Optimization.}
\begin{tabular}{@{}lcc@{}}
\toprule
Optimizer & Step wall (s), mean $\pm$ std & Peak GPU (MB), mean $\pm$ std \\
\midrule
SGD & $3.6 \times 10^{-1} \pm 5.9 \times 10^{-4}$ & $4.4 \times 10^{1} \pm 0.0$ \\
RMSprop & $3.4 \times 10^{-1} \pm 5.6 \times 10^{-4}$ & $4.4 \times 10^{1} \pm 0.0$ \\
Adam & $3.4 \times 10^{-1} \pm 1.1 \times 10^{-3}$ & $4.4 \times 10^{1} \pm 0.0$ \\
AdamW & $3.4 \times 10^{-1} \pm 1.4 \times 10^{-3}$ & $4.4 \times 10^{1} \pm 0.0$ \\
LBFGS & $2.0 \times 10^{0} \pm 1.6 \times 10^{-1}$ & $4.4 \times 10^{1} \pm 2.2 \times 10^{-2}$ \\
KFAC & $3.5 \times 10^{-1} \pm 2.1 \times 10^{-3}$ & $4.9 \times 10^{1} \pm 0.0$ \\
Shampoo & $3.4 \times 10^{-1} \pm 6.8 \times 10^{-4}$ & $4.6 \times 10^{1} \pm 0.0$ \\
SOAP & $3.4 \times 10^{-1} \pm 3.1 \times 10^{-3}$ & $4.6 \times 10^{1} \pm 0.0$ \\
RandomSearch & $4.3 \times 10^{0} \pm 7.2 \times 10^{-3}$ & $4.4 \times 10^{1} \pm 0.0$ \\
SimulatedAnnealing & $4.3 \times 10^{0} \pm 7.8 \times 10^{-3}$ & $4.4 \times 10^{1} \pm 0.0$ \\
ParticleSwarm & $2.9 \times 10^{0} \pm 2.2 \times 10^{-2}$ & $4.4 \times 10^{1} \pm 0.0$ \\
DifferentialEvolution & $3.0 \times 10^{0} \pm 5.0 \times 10^{-3}$ & $4.4 \times 10^{1} \pm 0.0$ \\
\bottomrule
\end{tabular}
\end{table}

\begin{table}[tbh]
\centering
\caption{Compute Resource Usage for Flight Dynamics Range Maximization.}
\begin{tabular}{@{}lcc@{}}
\toprule
Optimizer & Step wall (s), mean $\pm$ std & Peak GPU (MB), mean $\pm$ std \\
\midrule
SGD & $1.6 \times 10^{0} \pm 1.4 \times 10^{-2}$ & $1.1 \times 10^{1} \pm 0.0$ \\
RMSprop & $1.7 \times 10^{0} \pm 1.1 \times 10^{-2}$ & $1.1 \times 10^{1} \pm 0.0$ \\
Adam & $1.6 \times 10^{0} \pm 7.3 \times 10^{-3}$ & $1.1 \times 10^{1} \pm 0.0$ \\
AdamW & $1.5 \times 10^{0} \pm 5.0 \times 10^{-3}$ & $1.1 \times 10^{1} \pm 0.0$ \\
LBFGS & $5.0 \times 10^{0} \pm 4.9 \times 10^{0}$ & $1.1 \times 10^{1} \pm 2.1 \times 10^{-2}$ \\
KFAC & $2.1 \times 10^{0} \pm 5.1 \times 10^{-3}$ & $1.3 \times 10^{1} \pm 0.0$ \\
Shampoo & $2.1 \times 10^{0} \pm 1.8 \times 10^{-3}$ & $1.1 \times 10^{1} \pm 0.0$ \\
SOAP & $2.1 \times 10^{0} \pm 9.8 \times 10^{-3}$ & $1.1 \times 10^{1} \pm 0.0$ \\
RandomSearch & $1.7 \times 10^{1} \pm 6.4 \times 10^{-2}$ & $1.1 \times 10^{1} \pm 0.0$ \\
SimulatedAnnealing & $1.7 \times 10^{1} \pm 2.4 \times 10^{-2}$ & $1.1 \times 10^{1} \pm 0.0$ \\
ParticleSwarm & $1.1 \times 10^{1} \pm 1.1 \times 10^{-1}$ & $1.1 \times 10^{1} \pm 0.0$ \\
DifferentialEvolution & $7.6 \times 10^{0} \pm 7.6 \times 10^{-2}$ & $1.1 \times 10^{1} \pm 0.0$ \\
\bottomrule
\end{tabular}
\end{table}

\begin{table}[tbh]
\centering
\caption{Compute Resource Usage for Rigid-Body Robot Setpoint.}
\begin{tabular}{@{}lcc@{}}
\toprule
Optimizer & Step wall (s), mean $\pm$ std & Peak GPU (MB), mean $\pm$ std \\
\midrule
SGD & $2.9 \times 10^{0} \pm 5.4 \times 10^{-2}$ & $1.9 \times 10^{1} \pm 0.0$ \\
RMSprop & $2.8 \times 10^{0} \pm 2.2 \times 10^{-2}$ & $1.9 \times 10^{1} \pm 0.0$ \\
Adam & $2.6 \times 10^{0} \pm 3.5 \times 10^{-2}$ & $1.9 \times 10^{1} \pm 0.0$ \\
AdamW & $2.8 \times 10^{0} \pm 1.8 \times 10^{-1}$ & $1.9 \times 10^{1} \pm 0.0$ \\
LBFGS & $1.3 \times 10^{1} \pm 7.2 \times 10^{0}$ & $1.9 \times 10^{1} \pm 1.3 \times 10^{-2}$ \\
KFAC & $3.6 \times 10^{0} \pm 5.2 \times 10^{-3}$ & $1.9 \times 10^{1} \pm 0.0$ \\
Shampoo & $3.6 \times 10^{0} \pm 2.7 \times 10^{-2}$ & $1.9 \times 10^{1} \pm 0.0$ \\
SOAP & $3.7 \times 10^{0} \pm 5.6 \times 10^{-2}$ & $1.9 \times 10^{1} \pm 0.0$ \\
RandomSearch & $6.7 \times 10^{1} \pm 1.3 \times 10^{0}$ & $1.9 \times 10^{1} \pm 0.0$ \\
SimulatedAnnealing & $2.8 \times 10^{1} \pm 4.9 \times 10^{-2}$ & $1.9 \times 10^{1} \pm 0.0$ \\
ParticleSwarm & $1.4 \times 10^{1} \pm 5.0 \times 10^{-2}$ & $1.9 \times 10^{1} \pm 0.0$ \\
DifferentialEvolution & $5.7 \times 10^{0} \pm 1.5 \times 10^{-1}$ & $1.9 \times 10^{1} \pm 0.0$ \\
\bottomrule
\end{tabular}
\end{table}

\begin{table}[tbh]
\centering
\caption{Compute Resource Usage for Mass-Spring Robot Reach.}
\begin{tabular}{@{}lcc@{}}
\toprule
Optimizer & Step wall (s), mean $\pm$ std & Peak GPU (MB), mean $\pm$ std \\
\midrule
SGD & $1.4 \times 10^{0} \pm 1.1 \times 10^{-2}$ & $9.3 \times 10^{0} \pm 0.0$ \\
RMSprop & $1.4 \times 10^{0} \pm 2.3 \times 10^{-3}$ & $9.3 \times 10^{0} \pm 0.0$ \\
Adam & $1.4 \times 10^{0} \pm 3.8 \times 10^{-3}$ & $9.3 \times 10^{0} \pm 0.0$ \\
AdamW & $1.4 \times 10^{0} \pm 1.5 \times 10^{-2}$ & $9.3 \times 10^{0} \pm 0.0$ \\
LBFGS & $3.9 \times 10^{0} \pm 6.5 \times 10^{-1}$ & $9.4 \times 10^{0} \pm 2.3 \times 10^{-2}$ \\
KFAC & $7.5 \times 10^{-1} \pm 2.0 \times 10^{-3}$ & $1.1 \times 10^{1} \pm 0.0$ \\
Shampoo & $7.5 \times 10^{-1} \pm 4.0 \times 10^{-3}$ & $9.3 \times 10^{0} \pm 0.0$ \\
SOAP & $7.5 \times 10^{-1} \pm 9.1 \times 10^{-4}$ & $9.4 \times 10^{0} \pm 0.0$ \\
RandomSearch & $1.6 \times 10^{1} \pm 4.7 \times 10^{-2}$ & $9.3 \times 10^{0} \pm 0.0$ \\
SimulatedAnnealing & $1.6 \times 10^{1} \pm 7.2 \times 10^{-2}$ & $9.3 \times 10^{0} \pm 0.0$ \\
ParticleSwarm & $6.9 \times 10^{0} \pm 3.4 \times 10^{-2}$ & $1.0 \times 10^{1} \pm 0.0$ \\
DifferentialEvolution & $7.1 \times 10^{0} \pm 3.1 \times 10^{-2}$ & $9.4 \times 10^{0} \pm 0.0$ \\
\bottomrule
\end{tabular}
\end{table}

\end{document}